\documentclass[nohyperref]{article}

\usepackage{microtype}
\usepackage{graphicx}
\usepackage{subfig}
\usepackage{booktabs} 

\usepackage{pifont}
\usepackage{multirow}
\usepackage{threeparttable}
\usepackage{color, colortbl} 
\definecolor{Gray}{gray}{0.9}

\usepackage{hyperref}

\usepackage[accepted]{icml2022}

\usepackage{amsmath}
\usepackage{amssymb}
\usepackage{mathtools}
\usepackage{amsthm}

\usepackage[capitalize,noabbrev]{cleveref}

\theoremstyle{plain}

\theoremstyle{definition}

\theoremstyle{remark}

\usepackage[textsize=tiny]{todonotes}

\icmltitlerunning{\textit{Eshraghian and Lu}, The fine line between dead neurons and sparsity in BSNNs}

\begin{document}

\twocolumn[
\icmltitle{The fine line between dead neurons and sparsity\\
            in binarized spiking neural networks}




\begin{icmlauthorlist}
\icmlauthor{Jason K. Eshraghian}{umich,uwa}
\icmlauthor{Wei D. Lu}{umich}


\end{icmlauthorlist}

\icmlaffiliation{umich}{Department of Electrical Engineering and Computer Science, University of Michigan, MI, USA}
\icmlaffiliation{uwa}{Department of Computer Science and Software Engineering, University of Western Australia, WA, Australia}

\icmlcorrespondingauthor{Jason K. Eshraghian}{jasonesh@umich.edu}
\icmlcorrespondingauthor{Wei D. Lu}{wluee@umich.edu}

\icmlkeywords{Machine Learning, ICML}

\vskip 0.3in
]



\printAffiliationsAndNotice{}  


\begin{abstract}
Spiking neural networks can compensate for quantization error by encoding information either in the temporal domain, or by processing discretized quantities in hidden states of higher precision. In theory, a wide dynamic range state-space enables multiple binarized inputs to be accumulated together, thus improving the representational capacity of individual neurons. This may be achieved by increasing the firing threshold, but make it too high and sparse spike activity turns into no spike emission. In this paper, we propose the use of `threshold annealing' as a warm-up method for firing thresholds. We show it enables the propagation of spikes across multiple layers where neurons would otherwise cease to fire, and in doing so, achieve highly competitive results on four diverse datasets, despite using binarized weights. Source code is available at \url{https://github.com/jeshraghian/snn-tha/}.


\end{abstract}

\section{Introduction} \label{submission}
Memory access and data communication remain amongst the most expensive instructions in deep learning workloads 
\cite{hashemi2018learning,kozyrakis2010server}. To complement Feynman\footnote{And his American Physical Society lecture titled `There is Plenty of Room at the Bottom' \cite{feynman1959plenty}.}, there is plenty of room for optimization at the top of the stack: learning algorithms can be tweaked to decrease memory access frequency using activation sparsity, and to reduce memory usage by weight quantization.

\textbf{Activation sparsity:} spiking neural networks (SNNs) are an example of models that leverage activation sparsity ${\pmb{z} \in \{0, 1\}^n}$, where an activation ${z^i = 0}$ allows input-weight-multiply steps to be bypassed as the product is guaranteed to also return `0'. Memory access to read out weights can be skipped \cite{roy2019towards, eshraghian2021training}.

\textbf{Reduced weight precision:} parameter quantization directly reduces the number of memory cells that must be read at any compute cycle. In the extreme, binarized neural networks (BNNs) constrain each weight to $w^{ij} \in \{-1, 1\}$ at the cost of reduced representational capacity and challenging learning convergence \cite{hubara2016binarized, courbariaux2016binarized}. 


    

\begin{figure*}[!t] 
\centering
\subfloat[]
{
	\includegraphics[scale=0.22]{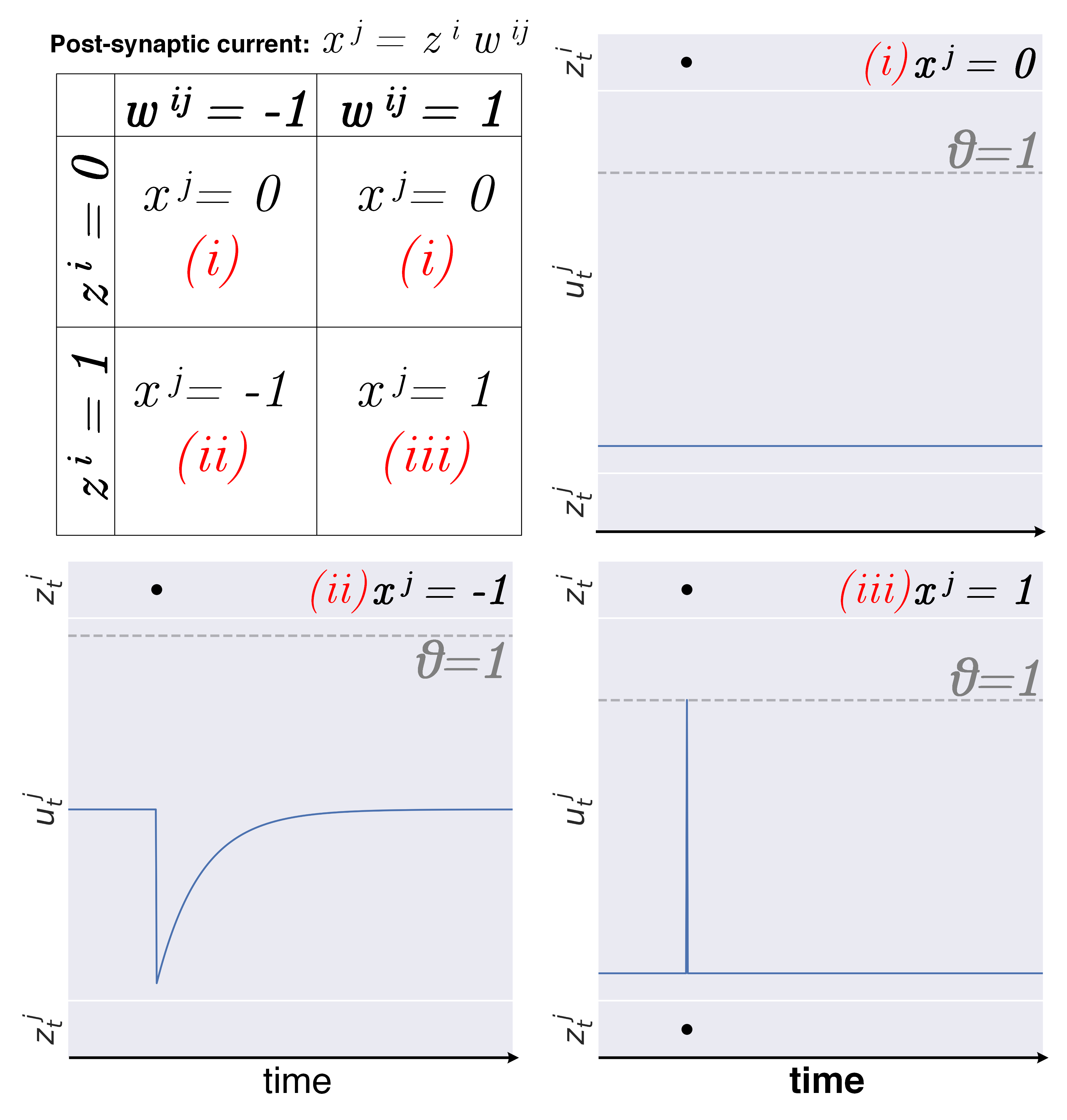}
	\label{fig1a}
}
\subfloat[]
{
	\raisebox{3.8em}{\includegraphics[scale=0.25]{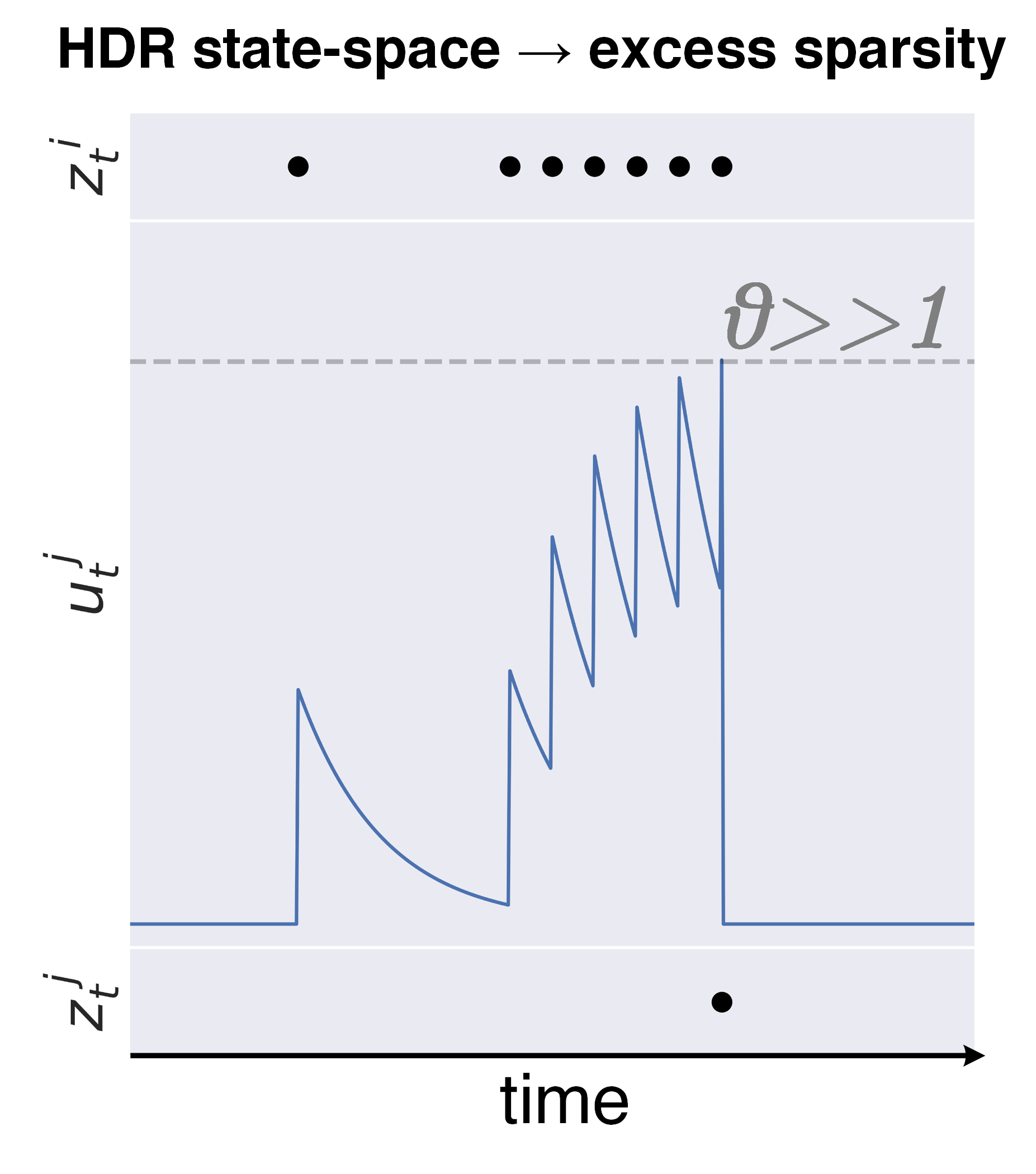}}
	\label{fig1b}
}
\subfloat[]
{
	\raisebox{3.8em}{\includegraphics[scale=0.25]{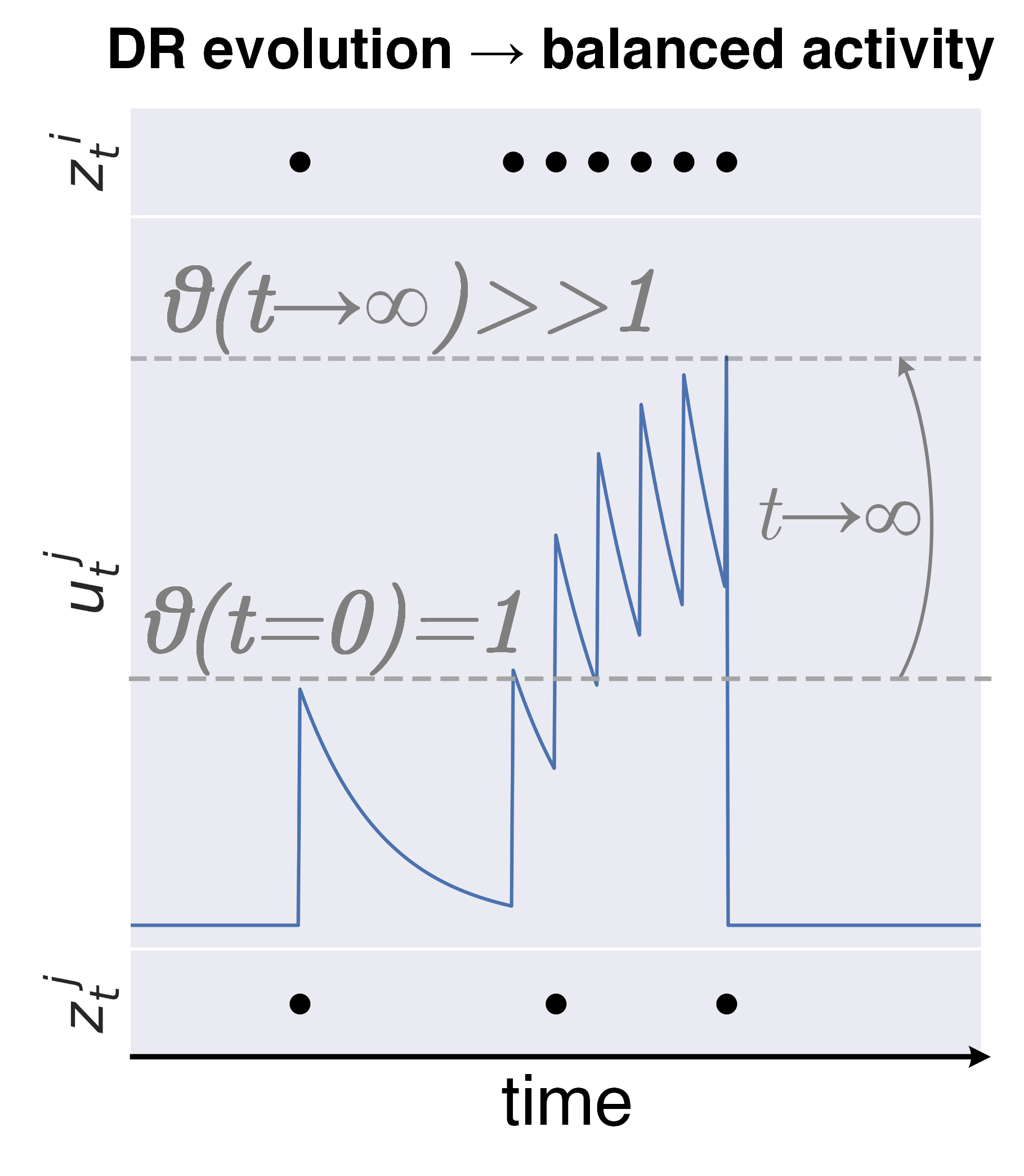}}
	\label{fig1b}
}
\caption{(a) Neuron response to the three possible post-synaptic currents $x^j$: (i) $x^j=0$: No response if $w^{ij}=0$, (ii) $x^j=-1$: Negative deflection with short-term memory but no spiking activity, (iii) $x^j=+1$: Spike propagation but no short-term memory due to reset. (b) High dynamic range (HDR) state precision comes at the expense of suppressed network activity. (c) Gradual evolution of the dynamic range (DR) balances precision and spike emission. Threshold annealing follows a similar principle, but the threshold $\theta$ is increased over training iterations instead of time. Spiking neuron parameters are thus fixed at run time, reducing computation and memory costs.}
\label{fig:1}
\end{figure*}

SNNs can amortize the cost of binarization by: 1) distributing information in the temporal domain rather than the activation amplitude \cite{voelker2020spike, kheradpisheh2021bs4nn}, and 2) accumulating binarized data in the high-precision, dynamically evolving state-space of spiking neurons \cite{schaefer2020quantizing, laborieux2021synaptic}. In both cases, discrete activations are applied to a continuous domain (either time or neuron state).

The dynamic range of the neuron state can be increased by using a large firing threshold. This is crucial for binarized SNNs (BSNNs), as it extends the range over which discrete inputs may be accumulated. However, indefinitely increasing the threshold impedes neuronal activity (and therefore, learning), leading to the `dead neuron problem'. To offset this issue, we propose `threshold annealing', a warm-up technique that improves classification accuracy in supervised learning tasks across four datasets (two static, two dynamic). This is achieved by dynamically increasing the threshold outside of the training-loop which enables:
\begin{itemize}
\item \textit{early} epochs to achieve sufficient network activity for low-precision learning to take place (i.e., by using a small threshold to promote adequate neuronal firing),
\item \textit{later} epochs to utilize the wide dynamic range state-space of spiking neurons for fine-tuning at increased precision (i.e., large thresholds that enable accumulation of multiple binarized quantities).
\end{itemize}

Our findings show state-of-the-art performance on all four datasets when compared to similarly lightweight networks, and a reduction of dead neurons on the most challenging dataset by 71\% when compared to BSNNs without threshold annealing, leading to faster training convergence.

\section{Background}
\subsection{Spiking Neuron Model}
SNNs adopt the same topologies modeled by conventional neural networks, but with the artificial neuron traded for a spiking neuron model. A single-state leaky integrate-and-fire neuron derived using the forward Euler method (\Cref{app:a1}) governed by the following discrete-time dynamics is used \cite{lapicque1907louis, gerstner2002spiking}:

\begin{equation}\label{eq:lif}
     u^j_{t+1} = \beta u^j_t + \sum_iw^{ij}z^i_{t+1} - z_{t}^j\theta,
\end{equation}

\begin{equation} \label{eq:spk}
    z^j_{t} =
    \begin{cases}
      1,  & \text{\rm if $u^j_t \geq \theta$} \\ 
      0, & \text{otherwise}, \\
    \end{cases}  
\end{equation}

where $u^j_{t}$ is the hidden state (membrane potential) of neuron $j$ at time $t$; $\beta$ is the decay rate of membrane potential; $w^{ij}$ is the weight between neurons $i$ and $j$; and a spike $z_t^j$ is elicited if the potential exceeds the threshold $\theta$. The final term of \Cref{eq:lif} resets the state by subtracting the threshold $\theta$ each time an output spike is generated.

\subsection{Binarized SNNs}\label{sec:BSNNs}

In BSNNs, each input spike $z^i_{t+1} \in \{0,1\}$ is scaled by a binarized weight $w^{ij} \in \{-1, 1\}$, resulting in a trinary post-synaptic potential $x^j_{t+1} = w^{ij}z_{t+1}^i \in \{-1, 0, 1\}$. This is added to the hidden state of a downstream neuron $u^j_t$, which triggers its own spike given sufficient excitation $u^j \geq \theta$. With threshold normalization,\footnote{Normalizing $\theta=1$ is common practice as it both enables the surrogate gradient term $\partial z^j/\partial u^j$ to be interpreted as a firing probability \cite{shrestha2018slayer, zenke2021remarkable}, and is also simple \cite{booij2005gradient}.} the three possible outcomes while the membrane potential is at rest are depicted in \Cref{fig:1}(a). In case (i), a nulled activation means no state change.  In case (ii), the neuron is inhibited and the hidden state follows a continuous trajectory back to its resting value. This is the only case where memory dynamics persist. In case (iii), the state immediately reaches the firing threshold $u^j = \theta$ upon spike arrival, and is subsequently forced to reset $u^j \leftarrow 0$. This is the only case where information is propagated to deeper layers. A problem emerges: short-term memory and spike emission are mutually exclusive. 

Using a large threshold is the most obvious solution \cite{lu2020exploring}, and achieves several benefits illustrated in \Cref{fig:1}(b):
\begin{itemize}
    \item the dynamic range of the state-space is increased,
    \item multiple inputs may be accumulated, and
    \item excitatory inputs can now be sustained as short-term, recurrent memory.
\end{itemize}

But these benefits do not indefinitely hold - too high a threshold obstructs the state from ever reaching it, causing spiking sparsity to extinguish into dead neurons. Large thresholds also increase susceptibility to vanishing gradients from an effect that is independent of dead neurons; surrogate gradients are typical activation functions (e.g., sigmoid, tanh, fast-sigmoid) centered about the firing threshold. Shift the surrogate function too far, and many neurons will dominantly operate in the subthreshold regime where $\partial z^i / \partial u^i \rightarrow 0$. 

Another drawback of using large thresholds is that it may interfere with gradient-based methods that rely on low thresholds to promote more neuronal activity, which is often the case for spike-time derivative methods and single-spike training schemes \cite{bohte2002error, booij2005gradient, xu2013supervised}. These conflicting trade-offs must somehow be balanced.

\subsection{Related Work}
Inspired by mechanisms that promote stability in the primary sensory cortex, threshold adaptation has been incorporated into computational models of synaptic plasticity \cite{bienenstock1982theory, zhang2003other}. Adaptation has been used as a homeostatic mechanism to regulate network activity in deep learning-inspired SNNs trained via local learning rules \cite{diehl2015unsupervised} and gradient-based optimization \cite{bellec2018long, shaban2021adaptive}. In all previous cases, threshold adaptation evolves over the course of simulated time steps (\Cref{fig:1}(c)). It has shown to be a potential approach to retaining long-term memory dynamics in SNNs using biologically plausible means, with the cost of each neuron updating its threshold trajectory at each time step. While this scales memory complexity linearly with the number of neurons $\mathcal{O}(n)$ during the forward-pass, the threshold must be recorded at every time step during training to implement the backpropagation (BPTT) algorithm, which additionally scales memory usage with time as well $\mathcal{O}(nT)$ 
\cite{rumelhart1986learning, werbos1990backpropagation}. This is overhead that we aim to eliminate in a compute and memory bound system, such as a BSNN, by i) moving the threshold evolution outside of the weight update loop, and ii) using a shared threshold for each layer. 

By eliminating the temporal variation of the threshold (i.e., it is only varied between training iterations), the purposes of membrane potential and threshold can be better decoupled. The membrane potential can be thought of as the mechanism for temporal data retention, while threshold warm-up leads to an evolving dynamic range state-space which enables inter-layer spike propagation.

\section{Threshold Annealing}

\begin{algorithm}[tb]
   \caption{Threshold Annealing}
   \label{alg:tha}
\begin{algorithmic}
   \REQUIRE $\theta_0 \in \mathbb{R}_{>0}$: Initial threshold
   \REQUIRE $\theta_\infty \in \mathbb{R}_{>\theta_0}$: Final threshold
   \REQUIRE $\alpha_\theta \in \mathbb{R}_{>0}:$ Inverse threshold time constant
   \REQUIRE $w:$ Initial parameters
   \REQUIRE $\eta:$ Learning rate
   \STATE $\theta_\gamma \leftarrow \theta_0$ (Initialize threshold)
   \STATE $s \leftarrow \partial \Tilde{z}(\theta_0)/\partial u$ (Initialize surrogate gradient using $\theta_0$)
   \STATE $\gamma \leftarrow 0$ (Initialize training step)
   \WHILE {$w$ not converged}
   \STATE $\gamma \leftarrow \gamma + 1$
   \STATE $g_\gamma \leftarrow \nabla_w f_\gamma(s, w_{\gamma-1})$ (Get gradients w.r.t. surrogate gradient $s$ using initial threshold $\theta_0$)
   \STATE $w_\gamma \leftarrow w_{\gamma - 1} - \eta \cdot g_\gamma$ (Update parameters e.g. using SGD or Adam)
   \STATE $\theta_\gamma \leftarrow \theta_{\gamma-1} + \alpha(\theta_\infty - \theta_{\gamma-1})$ (Update threshold without updating $s$)
    \ENDWHILE
    \STATE \textbf{return: }$w_\gamma$ (Resulting parameters) 
\end{algorithmic}
\end{algorithm}

\textbf{Out-of-the-loop threshold updates reduces complexity:} See \Cref{alg:tha} for pseudo-code of our proposed \textit{threshold annealing} method. \Cref{app:alg} provides more detailed pseudo-code that highlights how temporal evolution is isolated from spiking dynamics. A neuron's firing threshold exponentially relaxes from an initial value $\theta_0$ to a larger final value $\theta_\infty$ with time constant $\tau_\theta$:

\begin{equation} \label{eq:tha1}
    \tau_\theta \frac{d\theta_\gamma}{d\gamma} = -\theta_\gamma + \theta_\infty, \quad \theta_\infty > \theta_\gamma 
\end{equation}

where all neurons in each layer share the following explicit update rule at each training iteration $\gamma$:
\begin{equation}\label{eq:tha2}
\theta_{\gamma+1} = \theta_\gamma + \alpha(\theta_\infty - \theta_\gamma)    
\end{equation}

and $\alpha = \text{exp}(-1/\tau_\theta)$ is the inverse time constant of the threshold, noting that time here refers to the span of training iterations rather than the temporal dynamics of a spiking neuron. By only updating the threshold outside of the temporal evolution of each forward-pass (i.e., $\theta_\gamma$ is fixed during inference), memory no longer scales with time, and it also does not scale with iterations $\gamma$ as the BPTT algorithm truncates the computational graph between iterations. For shared thresholds per layer, memory scales with $\mathcal{O}(L) << \mathcal{O}(nT)$, where $L$ is the number of layers in the network.


\textbf{Fast to slow evolution balances weight updates with threshold updates:} The threshold dynamics are designed to follow the evolution of the effective learning rate. For example, when using the Adam optimizer, the estimation of moments (and therefore, weight updates) exponentially decays over iterations \cite{kingma2014adam}. 
Aligning the rate of threshold and weight updates is desirable as it avoids either of the two having an overpowering effect on the network: excessively large threshold updates will clip too many spikes at the start of training (similarly to \Cref{fig:1}(b)), whereas exceedingly small threshold updates fails to sufficiently extend the dynamic range of the state-space (the problems depicted in \Cref{fig:1}(a)). 




To ensure that a threshold update does not lead to the elimination of any spikes whilst still expanding the state-space of a neuron, the threshold change $\Delta \theta_\gamma = \theta_{\gamma + 1} - \theta_{\gamma}$ should be bounded by the distance between the new membrane potential at iteration $\gamma+1$ from the original threshold at $\gamma$:

\begin{equation}\label{eq:bound}
    \Delta \theta_\gamma < u_{\gamma + 1} - \theta_\gamma
\end{equation}

\begin{figure}[!t]
\begin{center}
\centerline{\includegraphics[scale=0.3]{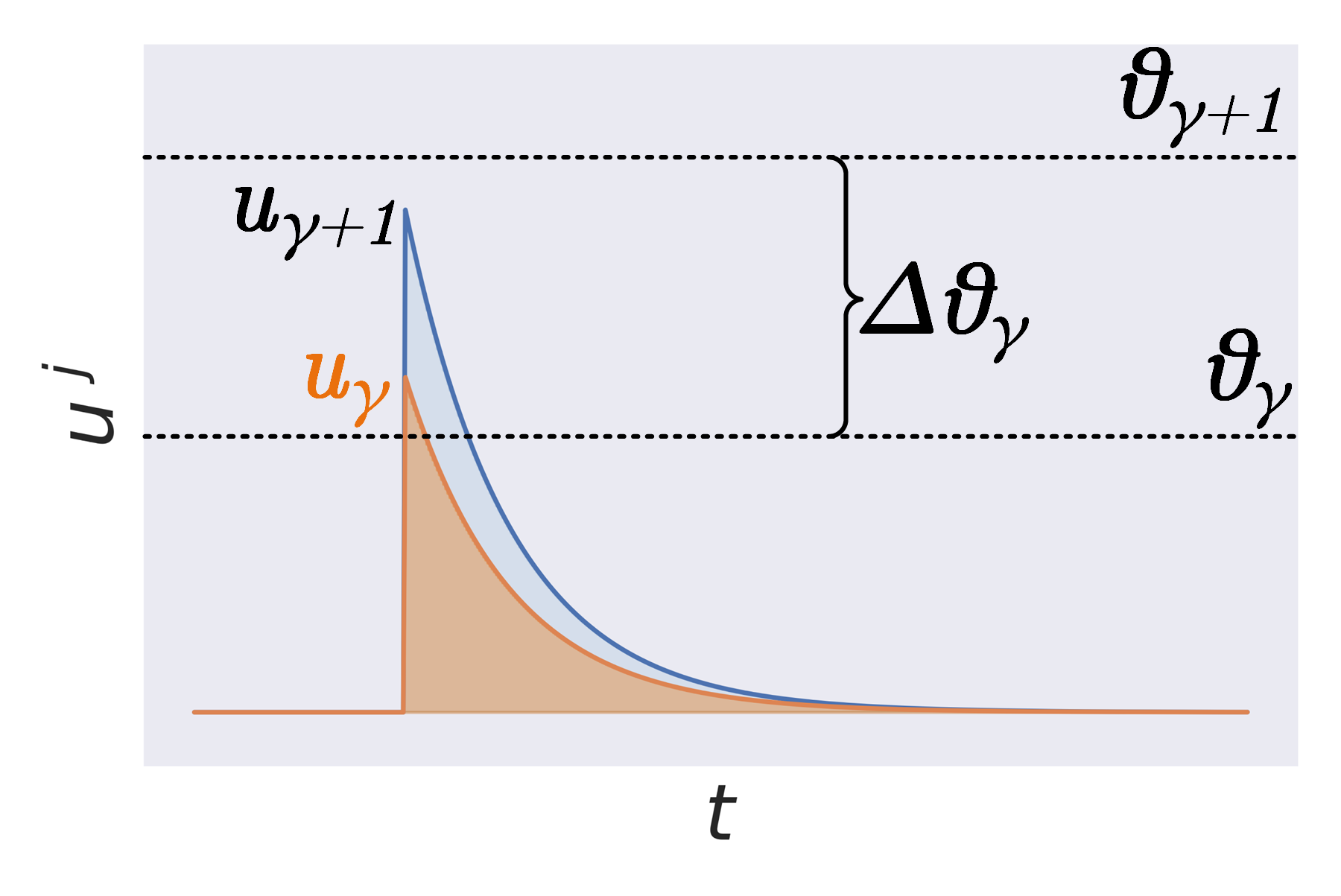}}
\caption{Membrane potential $u^j$ changes across iterations as a function of weight updates. If the threshold update across training iterations $\Delta \theta_\gamma$ is greater than the change of membrane potential, then a spike from iteration $\gamma$ will no longer occur at $\gamma+1$. Imposing the constraint from \Cref{eq:bound} prevents this. Note: the reset mechanism is not depicted for visual clarity.}
\label{fig:evo}
\end{center}
\end{figure}

This is visually depicted in \Cref{fig:evo}, where neglecting the condition from \Cref{eq:bound} would suppress spikes. If this condition is neglected for all of time $t$, the neuron would cease to contribute to the loss function, and lack a gradient signal to contribute to learning, i.e., a dead neuron. From the derivation provided in \Cref{app:a3}, the following two constraints can be imposed on $\theta_\gamma$ and $\theta_\infty$ to prevent threshold annealing from contributing to dead neurons:


\begin{equation}\label{eq:rate1}
    (1-\alpha)\theta_\gamma + \alpha\theta_\infty < \mathbf{z} \cdot~ \Big(\mathbf{w}_{\gamma}^{~:, j} + \eta \frac{\partial \mathcal {L}}{\partial \mathbf{w}^{~:,j}_\gamma}\Big)
\end{equation}

\begin{equation}\label{eq:rate2}
        \theta_\infty < \mathbf{z} \cdot~ \mathbf{w}_{\infty}^{~:, j}
\end{equation}

where $\mathbf{z} \in \mathbb{R}^i$ is the input vector, and $\mathbf{w}_{\gamma}^{~:, j} \in \mathbb{R}^i$ is the weight vector connected to the $j^{th}$ neuron of the first layer, and $\mathbf{w}_{\infty}^{~:, j}$ is the final weight vector at the end of training.

The weight update term on the right-hand side of \Cref{eq:rate1} assumes stochastic gradient descent (SGD) is used, where for adaptive optimizers the derivative term often decays exponentially with network converges. As $\theta_\infty$ is a constant term, the evolution of $\theta_\gamma$ should also exponentially diminish. This is adhered to in \Cref{eq:tha2}, and balances the two objectives of i) using a large threshold, while ii) not inducing dead neurons due to threshold updates. 

More intuitively, during early training iterations, the weight updates are large enough to `catch up' to the effect of a large threshold step. This is because the effect of weight decay is minimal during early training iterations. As the network is trained for more epochs, threshold updates are reduced to match that of weight decay.

In the experiments that follow, we adhere to exponential decay of threshold update magnitude as per \Cref{eq:tha2} and choose $\theta_\infty$, $\theta_\gamma$ and $\alpha$ based on a hyperparameter search. For sparsely populated input data, i.e., where the mean of the input features $\mathbf{z}$ over time is quite small (e.g., the Spiking Heidelberg Digits dataset \cite{cramer2020heidelberg}), the optimal value of $\alpha$ reduces by several orders of magnitude to slow down the rate at which $\theta_\gamma$ converges to $\theta_\infty$ (see \Cref{app:temporal}). This result is independently predicted by \Cref{eq:rate1}.



\begin{figure}[!t]
\begin{center}
\centerline{\includegraphics[scale=0.3]{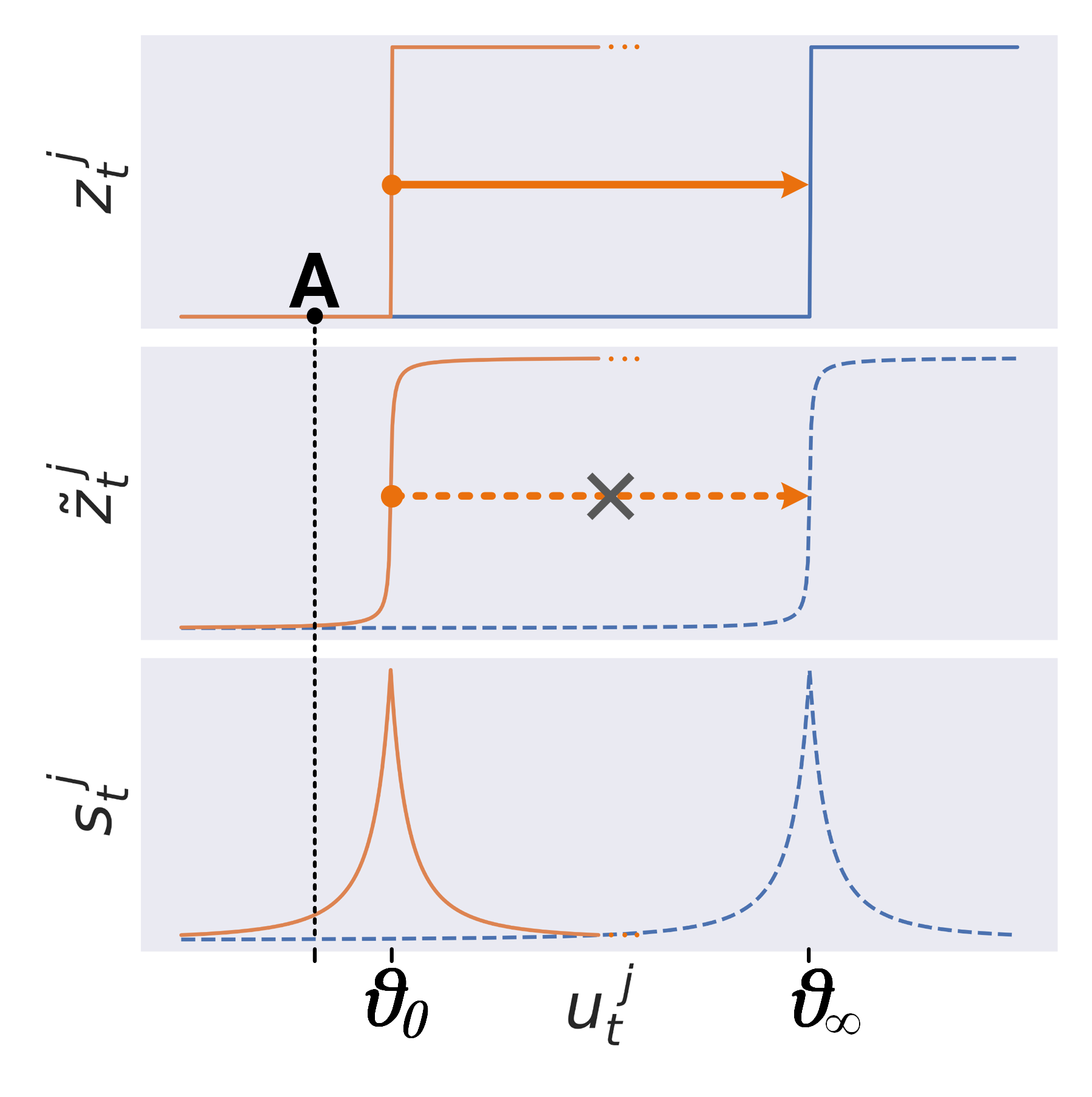}}\caption{\textbf{Top row:} output spiking activity $z^j_t$ is represented with the Heaviside operator centered about the firing threshold. Over iterations, threshold annealing pushes the Heaviside operator along the x-axis. \textbf{Middle row:} Surrogate of the Heaviside operator to address non-differentiability $\tilde{z}^j_t$. Our approach does not shift the surrogate function which reduces the impact of vanishing gradients on subthrehsold neurons (e.g., at point \textbf{A}). \textbf{Bottom row:} the surrogate gradient $s^j_t$. The intersection between a neuron at point \textbf{A} and the non-shifted gradient is larger than that of the shifted gradient, thus providing latent/quiet neurons with the opportunity to continue learning, and therefore contribute to network activity.}
\label{fig:grad}
\end{center}
\end{figure}

\textbf{Clamped surrogate gradients boost the weight update signal for subthreshold neurons: }
The conventional method to account for the non-differentiability of \Cref{eq:spk} is to use a surrogate gradient which substitutes the Heaviside operator with a threshold-shifted, differentiable alternative e.g., the sigmoid function, in the backward pass (see \Cref{fig:grad}). If this approach was strictly adhered to, then the surrogate function would shift along with the threshold over training iterations.

Consider a neuron with a subthreshold membrane potential at point \textbf{A}, depicted in \Cref{fig:grad}. Shifting the surrogate function along with the threshold causes $s^j_t \rightarrow 0$, i.e., the vanishing gradient problem (contrasted from the dead neuron problem, where neurons cease to fire). 


\begin{figure*}[!t] 
\centering
\subfloat[]
{
	\includegraphics[scale=0.27]{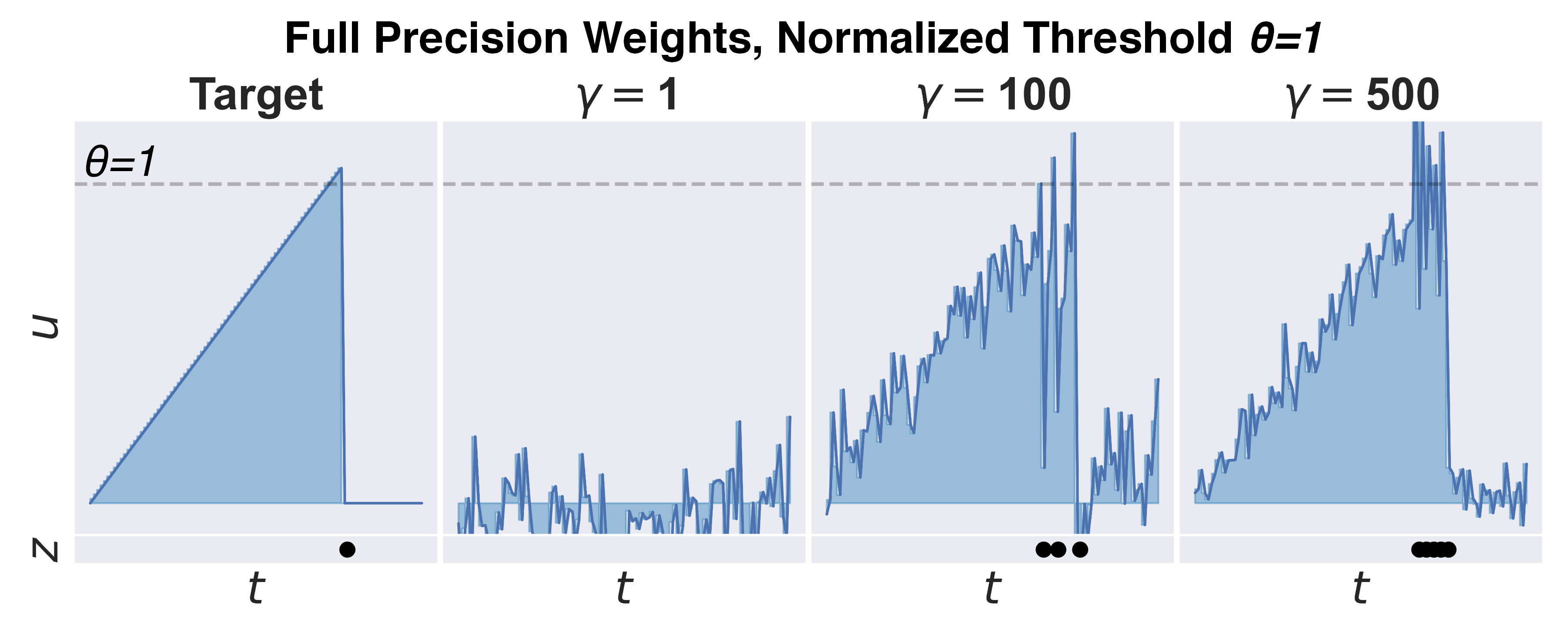}
	\label{fig4a}
}
\subfloat[]
{
\includegraphics[scale=0.27]{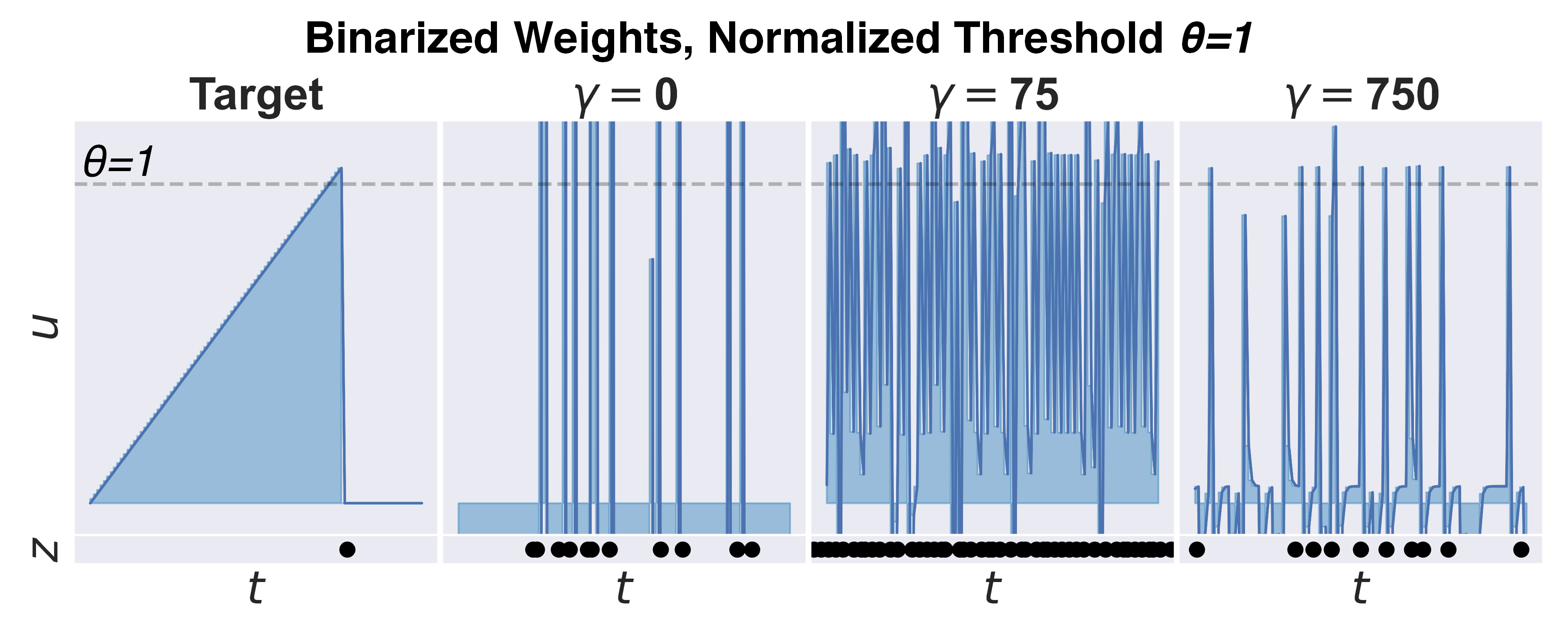}
	\label{fig4b}
}\\
\subfloat[]
{
\includegraphics[scale=0.27]{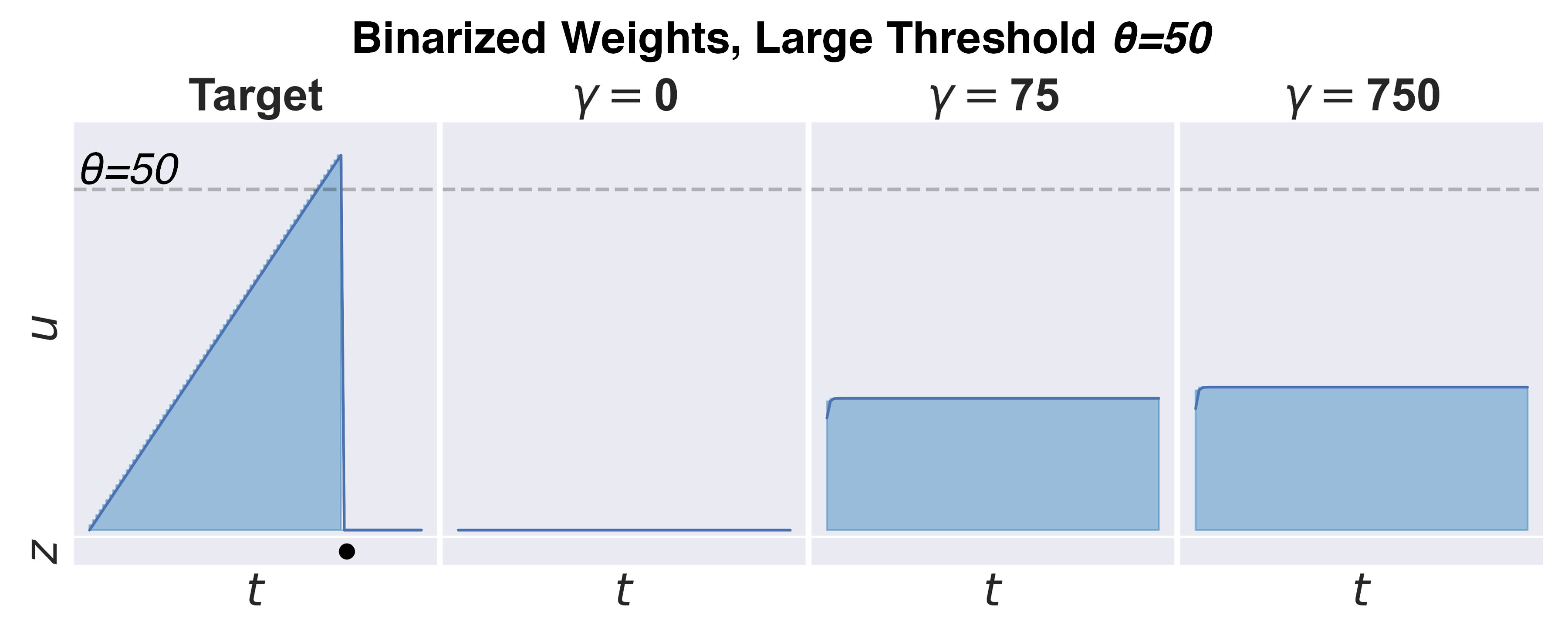}
	\label{fig4c}
}
\subfloat[]
{
\includegraphics[scale=0.27]{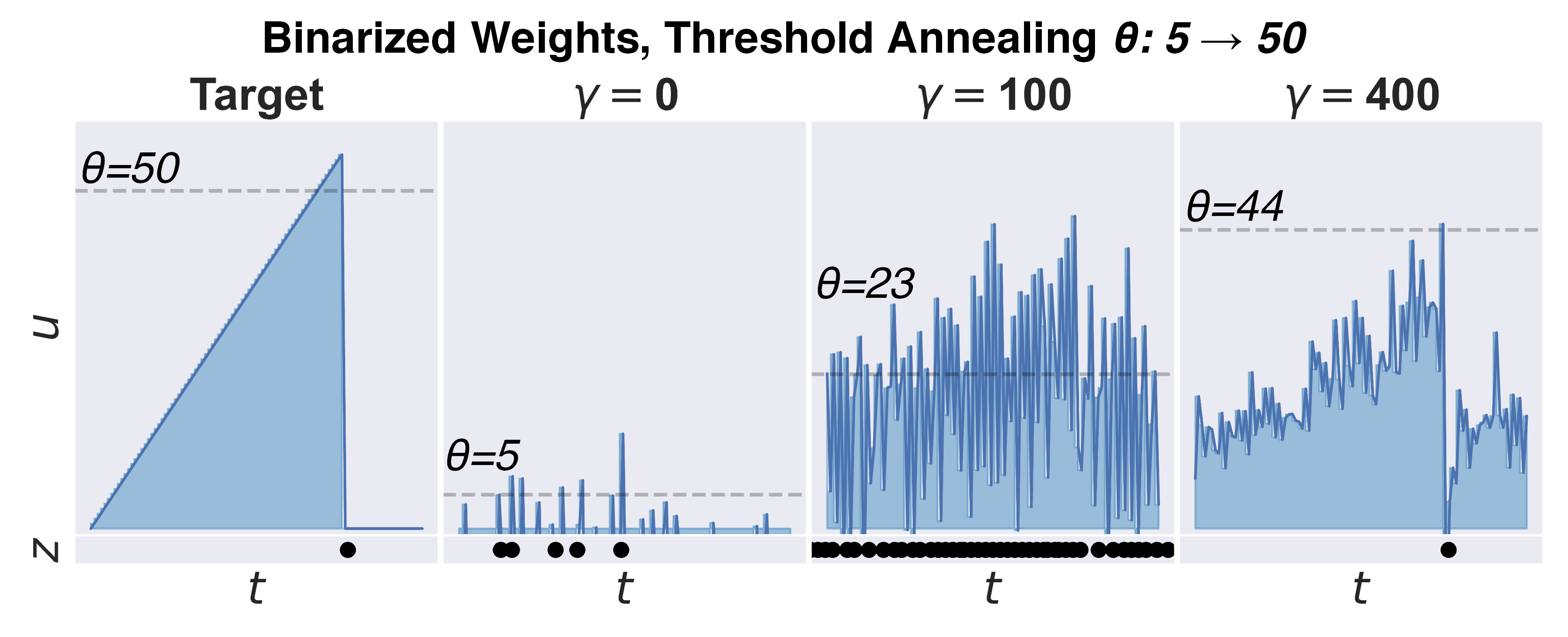}
	\label{fig4d}
}
\caption{An intuitive demonstration of threshold annealing on a spike timing task. (a) Full precision weights with threshold $\theta=1$. (b) Binarized weights with threshold $\theta=1$ leads to unstable neuronal activity. (c) Binarized weights with a large threshold $\theta=50$ suppresses all firing activity; the increase in hidden state $u$ is due to an increasing bias. (d) Binarized weights with threshold annealing from $\theta: 5 \rightarrow 50$ successfully learns to fire at the desired spike-time at the 400th iteration. The performance better matches that of the full precision network. Animated visualizations are provided at the following link: \url{github.com/jeshraghian/snn-tha/}.}
\label{fig:time}
\end{figure*}

To address this issue, our implementation clamps the surrogate gradient to the initial threshold value $\theta_0$, preventing the function from shifting during the training process. This is highlighted in \Cref{alg:tha}, where the threshold $\theta_\gamma$ is updated without updating the surrogate gradient $s$. Our experiments use the fast sigmoid function:

\begin{equation}\label{eq:surrogate}
    s^j_t = \frac{\partial z^j_t}{\partial u^j_t} \leftarrow \frac{1}{(1+k|\theta_0 - u^j_t|)^2}
\end{equation}

\noindent where the left arrow denotes substitution, and $k$ is a slope that modulates the linearity of the surrogate function. While the learning process is generally robust to the choice of the surrogate gradient \cite{zenke2021remarkable}, we choose to fix the gradient based on the initial threshold $\theta_0$ as it removes the need to re-compute a new surrogate function at each iteration.

The primary benefit is that the gradient signal is boosted for neurons closer to their resting potential. Quantitatively, for a neuron at rest $u_t^j \approx 0$ and $\theta >> 1$,  and assuming $k=1$, then $s_t^j \approx 1/\theta^2$ (based on \Cref{eq:surrogate}). In practice, $\theta_\infty$ can be close to one order of magnitude larger than $\theta_0$ (see \Cref{app:timing} and \cref{app:temporal} for examples). If $\theta_\gamma$ is used in \Cref{eq:surrogate} instead, this would reduce $s_t^j$ (and by extension, $\nabla_w\mathcal{L}$) by the square of this factor as $\theta_\gamma \rightarrow \theta_\infty$.

This benefit does not come for free, as active neurons in later training iterations (where $u^j_t \approx \theta_\infty$) face the same problem of a vanishing $s_t^j$. We claim this is tolerable for two empirical reasons: i) upon spike emission, the membrane potential of an active neuron is reset where a low $u^j$ provides an opportunity for the neuron to learn at another time step, and ii) the neuron must have already been trained in earlier iterations for its membrane potential to reach a large value $\theta_\infty$ (see \Cref{fig:time}(d) for an experiment where $u^j$ starts at rest and is trained to approach $\theta_\infty=50$). Quiet neurons are therefore provided with a better opportunity to contribute to network activity via gradient clamping.


\section{Experimental Results}
We test a full precision SNN and several BSNNs under the following cases i) with a normalized threshold $\theta=1$, ii) with a fixed threshold determined by a hyperparameter search, and iii) using threshold annealing, for each experiment on four datasets ranging from simple to increasingly difficult: MNIST \cite{lecun1998gradient}, FashionMNIST \cite{xiao2017fashion}, DVS128 Gesture \cite{amir2017low}, and Spiking Heidelberg Digits (SHD) \cite{cramer2020heidelberg}, without dataset augmentation. All experiments were performed on a 16GB NVIDIA V100 GPU, conducted across 5 trials each. Hyperparameters were selected based on a sweep over 500 trials for each individual experiments (see \Cref{app:exp} for further details). We used snnTorch 0.4.11 to construct spiking neuron models~\cite{eshraghian2021training} which uses a PyTorch 1.10.1 backend~\cite{paszke2019pytorch} in Python 3.7.9.

\subsection{Temporal Coding}\label{sec:TC}

Before moving to more complex data-driven tasks, we first provide an interpretable demonstration of how threshold annealing can improve BSNN performance on a straightforward spike-time encoding task. A single output neuron is trained to fire at a target time step by encouraging the membrane potential to linearly increase to the threshold at $t=75$, and subsequently remain quiet upon spike emission. A 3-layer BSNN is used with a fully-connected architecture of 100-1000-1 neurons. The input to the network is a Poisson spike train simulated across 100 time steps. The mean square error between the target and actual membrane potential are summed at each time step. Precise implementation details are provided in \Cref{app:timing}.

\Cref{fig:time} depicts the membrane potential trace of the output neuron over time at several training iterations $\gamma$. The full precision network in (a) accomplishes the task with ease, while the BSNN with $\theta=1$ in (b) highlights the challenges described in \Cref{sec:BSNNs}, i.e., the neuron is unable to both emit spikes as well as retain memory. Increasing the threshold of all neurons to $\theta=50$ completely suppresses spiking activity in the hidden and output layers. The membrane potential increases due to the bias, and settles at approximately the midpoint between steady-state and threshold, where the loss of the triangular membrane potential target is minimized. To fix the memory leakage problem in (b) and the dead neuron problem in (c), threshold annealing is applied by gradually warming up $\theta_0=5$ to $\theta_\infty=50$ with an inverse threshold time constant $\alpha=5\times10^{-3}$ (\Cref{eq:tha2}). The low threshold at the first iteration $\gamma=0$ enables spike propagation throughout layers, which sustains sufficient activity to keep the network learning over epochs. At the 400th iteration, the BSNN with threshold annealing successfully emits a spike at the target time step. The output neuron continued to remain active at the end of the simulation, where $\theta_\gamma$ was within 0.1\% of the $\theta_\infty$; a definitive improvement over use of a fixed threshold in \Cref{fig:time}(c). Animations in the source code offer further visual intuition on how threshold annealing `pulls up' neuronal activity.

\subsection{Static Datasets}\label{sec:static}
The MNIST~\cite{lecun1998gradient} and the slightly more challenging FashionMNIST datasets~\cite{xiao2017fashion} are used to assess the performance of threshold annealing on temporally static data. Both datasets consist of 60,000 28$\times$28 greyscale images in the training set, and 10,000 images in the test set, with 10 output classes of handwritten digits (MNIST) and clothing items/accessories (FashionMNIST). The convolutional BSNN architecture is provided in the \Cref{tab:static} footnote: 16Conv5 refers to a $5\times5$ convolutional kernel with $16$ filters; $AP2$ is a $2\times2$ average-pooling layer applied to membrane potential, and 1024Dense10 is a dense layer with 1024 inputs and 10 outputs.

The raw datasets are passed to the BSNN for 100 time steps of simulation without input encoding. The predicted class is set to the neuron with the highest spike count at the end of the 100 time steps. Hyperparameters and the optimization process for each experiment are provided in \Cref{app:static}, and the final results are shown in \Cref{tab:static}, where our proposed threshold annealing technique improves the performance of BSNNs by an average of 0.3\% for MNIST, and 0.97\% for FashionMNIST over the next best BSNN. The worst-case scenario of a BSNN with a normalized threshold still performs reasonably well, on account of static datasets not requiring any memory retention. Performance degrades further when temporal dynamics are included. 

\setlength{\tabcolsep}{5pt}
\newcolumntype{g}{>{\columncolor{Gray}}c}
\begin{table}[!t] 
\centering
\caption{Test set accuracies for static datasets.}\label{table:main_result}
\vskip 0.15in
\begin{threeparttable}
\begin{tabular}{lgggccc} \toprule \toprule \label{tab:static}
\multirow{2}{*}{} & \multicolumn{3}{c}{\textbf{MNIST}} & \multicolumn{3}{c}{\textbf{FashionMNIST}} \\

 & \multicolumn{1}{l}{\textbf{Best}} & \multicolumn{1}{l}{\textbf{Mean}} & \multicolumn{1}{l}{\textbf{~~~$\sigma$}} & \multicolumn{1}{l}{\textbf{Best}} & \multicolumn{1}{l}{\textbf{Mean}} & \multicolumn{1}{l}{\textbf{~~~$\sigma$}} \\
 \midrule
 
flt32 & 99.45 & 99.31 & 0.12 & 91.13 & 91.03 & 0.09 \\ \midrule

BSNN,$\theta$:1 & 98.46 & 98.44 & 0.02 & 87.11 & 86.70 & 0.36  \\
BSNN & 98.81 & 98.77 & 0.06 & 87.01 & 86.95 & 0.10  \\

\textbf{Proposed} & \textbf{99.12} & \textbf{99.07} & \textbf{0.04} & \textbf{88.12} & \textbf{87.92} & \textbf{0.22}   \\ \bottomrule \bottomrule

\end{tabular}
\begin{tablenotes}
    \item[1] 16Conv5-AP2-64Conv5-AP2-1024Dense10.
    \item[2] $\sigma$: Sample standard deviation.
\end{tablenotes}
\end{threeparttable}
\end{table}

\setlength{\tabcolsep}{5pt}
\newcolumntype{g}{>{\columncolor{Gray}}c}
\begin{table}[!t] 
\centering
\caption{Test set accuracies for temporal datasets.}\label{table:temporal}
\vskip 0.15in
\begin{threeparttable}
\begin{tabular}{lgggccc} \toprule \toprule
\multirow{2}{*}{} & \multicolumn{3}{c}{\textbf{DVS128 Gesture}} & \multicolumn{3}{c}{\textbf{SHD}} \\

 & \multicolumn{1}{l}{\textbf{Best}} & \multicolumn{1}{l}{\textbf{Mean}} & \multicolumn{1}{l}{\textbf{~~~$\sigma$}} & \multicolumn{1}{l}{\textbf{Best}} & \multicolumn{1}{l}{\textbf{Mean}} & \multicolumn{1}{l}{\textbf{~~~$\sigma$}} \\
 \midrule
 
flt32 & 93.05 & 92.87 & 0.32 & 82.51 & 82.27 & 0.27 \\ \midrule

BSNN,$\theta$:1 & 84.37 & 84.03 & 0.60 & 63.86 & 62.99 & 0.91  \\
BSNN & 89.24 & 88.54 & 0.70 & 62.94 & 62.48 & 0.41  \\

\textbf{Proposed} & \textbf{92.36} & \textbf{91.32} & \textbf{0.56} & \textbf{66.35} & \textbf{65.68} & \textbf{0.66}   \\ \bottomrule \bottomrule

\end{tabular}
\begin{tablenotes}
    \item[1] DVS128: 16Conv5-AP2-32Conv5-AP2-8800Dense11.
    \item[2] SHD: 700Dense3000Dense20.
    \item[3] $\sigma$: Sample standard deviation.
\end{tablenotes}
\end{threeparttable}
\end{table}

\subsection{Temporal Datasets}\label{sec:temporal}
The DVS128 Gesture dataset~\cite{amir2017low} is filmed with an event-based camera~\cite{patrick2008128x} which only processes sufficient changes in luminance, and consists of 11 different output classes of hand gestures, such as clapping, arm rotation, and air guitar. To fit the dataset within 100 time steps, events were integrated over a period of 5~ms such that each sample in the training and test set are approximately 1~s and 3.5~s in duration, respectively. Spatial downsampling fits each sample to dimensions of $(2\times32\times32)$. 

As a more challenging alternative, the SHD dataset was also used which consists of 20 output classes of audio recordings of the ten digits spoken aloud in both the English and German languages. The recordings are processed using a cochlear model for spike encoding \cite{cramer2020heidelberg}.

The final test accuracies are shown in \Cref{table:temporal}, with experimental and hyperparameter details in \Cref{app:temporal}. As dataset complexity increases, the performance of BSNNs degrade further, while threshold annealing offsets the degradation to a larger degree. For the DVS128 Gesture dataset, when compared to the full precision network, the baseline BSNNs with and without threshold normalization drop by 8.8\% and 4.4\%, respectively. But this performance drop is reduced significantly to 1.5\% when threshold annealing is used during training. These results indicate that BSNNs have an optimal solution that does not deviate far from the high precision case; the challenge is finding this solution. Methods such as threshold annealing make the training process easier.

Performance drops drastically for the SHD dataset when binarizing weights across all cases. While threshold annealing improves upon performance by 2.7\% compared to the next best BSNN, it still remains at a large distance from the high precision performance. While this may partially be accounted for by complexity of the SHD dataset, it may also be due to the additional hyperparameters that are introduced which require a larger number of trials during the hyperparameter search process. For fairness, the number of random trials was set to 500 for all networks, though threshold annealing would likely benefit from increasing the search space.

\section{Discussion} \label{sec:dis}

\begin{figure*}[!t] 
\centering
\subfloat[]
{
	\includegraphics[scale=0.22]{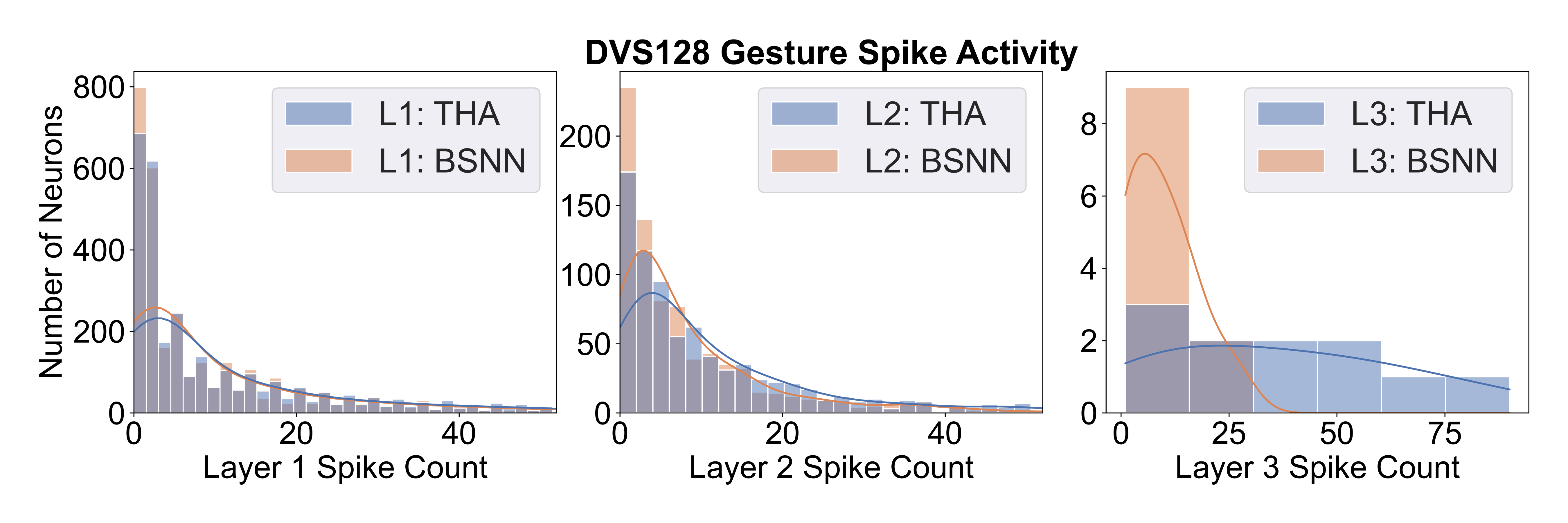}
	\label{fig5a}
}
\subfloat[]
{
\includegraphics[scale=0.245]{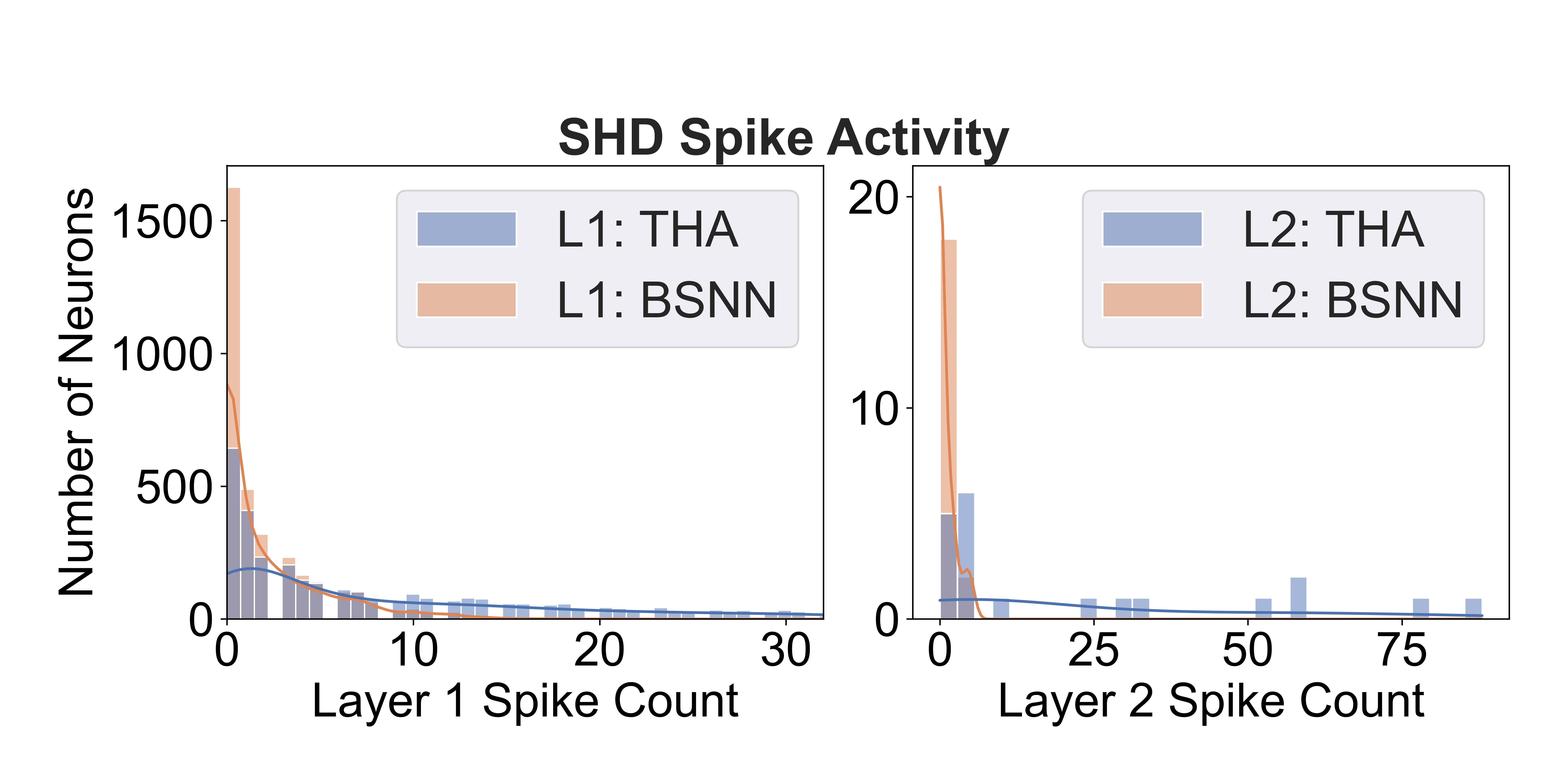}
	\label{fig:5b}
}\\

\caption{Spiking activity across layers with threshold annealing (THA) and without (BSNN) on temporal datasets. Static dataset distributions are provided in \Cref{app:dist}. (a) \textbf{DVS128 Gesture:} Distribution of spikes in the output layer with threshold annealing is balanced, whereas most neurons are silent without annealing. (c) \textbf{SHD:} Using a dataset with sparse activity results in a large number of dead without threshold annealing.}  
\label{fig:sparsity}
\end{figure*}

\begin{figure*}[!t] 
\centering
\includegraphics[scale=0.2]{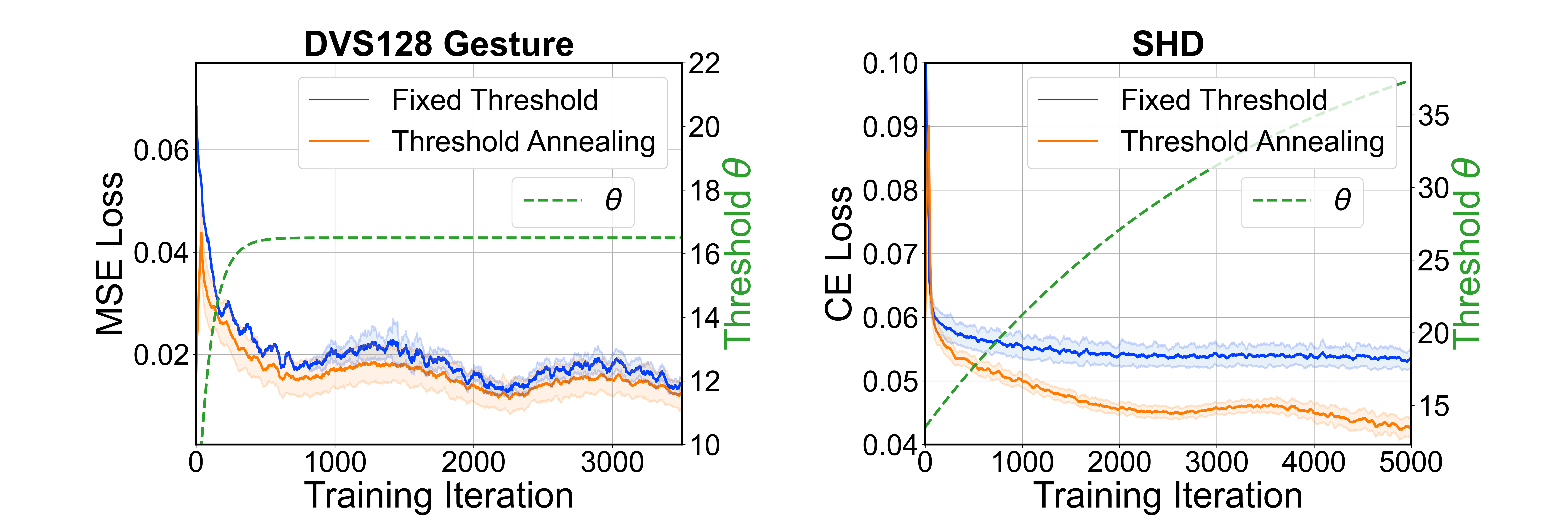}
\caption{Loss curves on dynamic datasets. Oscillations are a result of using periodic learning rate schedule.}
\label{fig:loss}
\end{figure*}

\subsection{Comparison with lightweight SNNs}
There is a general lack of reproducible baselines that implement gradient-based learning on convolutional BSNNs, likely due to hyperparameter sensitivity and the associated difficulty in training. The few related approaches include Lu and Sengupta's ANN-SNN conversion which is benchmarked on large-scale static datasets \cite{lu2020exploring}. Kheradpisheh \textit{et al.} use temporal backpropagation (BS4NN) on dense BSNNs \cite{kheradpisheh2021bs4nn}. We further consider several additional promising lightweight approaches to training full precision SNNs, including sparse spiking gradient descent (SSGD) which reduces overhead during gradient calculations \cite{perez2021sparse}, and neural heterogeneity (NH) which compresses the required number of neurons with the addition of neuron-independent parameters \cite{perez2021neural}. Interestingly, BSNNs with threshold annealing can outperform full precision SNNs in terms of accuracy using considerably less memory for model parameters. These baselines rely on dense networks only, which is why we developed our own full precision and BSNN baselines in previous sections.


\textbf{FashionMNIST:} SSGD and NH achieve 86.7\% and 87.5\%, respectively, using full precision weights. NH requires 101k full precision weights (3.2~Mbits). BS4NN achieves 87.3\% using 794~Kbits of binarized parameters. Threshold annealing outperforms in terms of accuracy at 87.9\% with far less memory footprint with 12k binarized weights (12~Kbits).

\textbf{DVS128 Gesture:} NH obtains 82.1\%, whereas our BSNN approach obtains 91.3\% with less memory overhead. Assuming the input dimensions of the DVS128 Gesture dataset are downsampled by a factor of 4, the network in NH requires 8.4~Mbits of full precision weights as against 98~Kbits of binarized weights in our approach.

\textbf{SHD:} In contrast to the above results, both full precision NH (82.7\%) and SSGD (77.5\%) outperform our BSNN (65.7\%). This can be attributed to the use of recurrent connections in NH and SSGD; without recurrent SNNs in SSGD, performance dropped below 50\%. Likewise from Cramer \textit{et al.}'s original proposal of the SHD dataset, the test accuracy was 48.6\% on a similarly 2-layer dense network \cite{cramer2020heidelberg}, which our BSNN with threshold annealing outperforms by around 17\%. When recurrent connections were added, test accuracy jumped to 83.2\%. Although we use a considerably wider hidden layer (3000 neurons vs 128 neurons in NH), binarized weights still occupy less memory (ours: 2.16~Mbits vs NH: 2.95~Mbits).

\subsection{Spiking Activity and Convergence Rate}
\Cref{fig:sparsity} shows the distribution of spiking activity across layers both with and without threshold annealing. This is measured by passing the entire test set to the network and taking the average spike count for each neuron. For the fixed threshold network trained on the SHD dataset, 55\% and 90\% of neurons in the first and final layer do not contribute to network activity, compared to approximately 20\% with threshold annealing in both layers.
This is highly undesirable at the start of the learning process as gradients cannot backpropagate through neurons that do not contribute to the final loss, which causes slower training convergence. This is shown in \Cref{fig:loss}, where the larger proportion of dead neurons in the fixed threshold network trained on the SHD dataset leads to worse performance. Averaging across both layers, threshold annealing reduces the number of dead neurons by 71\% in the SHD network. 

The threshold dynamics of the final layer of both networks are also overlayed, where the rate of increase is slower when trained on SHD than DVS128 Gesture. This is a result of sparser input activity, and was predicted by \Cref{eq:rate2}. Once the training process is terminated based on early stopping criteria, the same thresholds of the optimal trial should be used. In all experiments, the final threshold $\theta_\gamma$ was always within 0.1\% of the steady-state threshold $\theta_\infty$, so we simply use $\theta_\infty$ at test time.


\subsection{Concluding remarks}
In general, our approach to training BSNNs is extremely competitive when compared to high precision, lightweight networks that consume more memory. As our implementation is based on a PyTorch backend, the binarized weights are represented as full precision values and are not explicitly stored with compressed memory. Rather, the models generated are intended to be ported onto dedicated neuromorphic hardware that is tailored for single-bit weight representations \cite{orchard2021efficient, akbarzadeh2020digital, rahimi2020complementary}.

It is also important to account for how this memory is used. For example, while the memory overhead of synaptic weights scale with the number of neurons as $\mathcal{O}(n^2)$, SNNs in general are power efficient as this memory can be accessed with low frequency. On the other hand, neuron specific parameters, such as in neural heterogeneity \cite{perez2021neural} and threshold adaptation \cite{bellec2018long}, have less memory complexity $\mathcal{O}(n)$, but this memory must be accessed at each time step to update the hidden state and does not benefit from spike sparsity. By sharing thresholds for neurons in given layers, threshold annealing not only adds negligible memory footprint (i.e., one full precision value per layer), but it does so with good performance. Combining threshold annealing with heterogenous time constants and recurrent connections could attract the benefits of both worlds in lightweight BSNN computation, especially where intrinsic circuit-level variations can be exploited as diverse neuronal behavior \cite{kang2021build}.








\newpage
\section*{Acknowledgements}
This work was supported, in part, by the Semiconductor Research Corporation and DARPA, through the Applications Driving Architectures Research Center, and the National Science Foundation through awards ECCS-191550 and CCF-1900675. Jason K. Eshraghian is supported by the Forrest Research Fellowship through the Forrest Research Foundation.




\bibliography{example_paper}

\begin{thebibliography}{44}
\providecommand{\natexlab}[1]{#1}
\providecommand{\url}[1]{\texttt{#1}}
\expandafter\ifx\csname urlstyle\endcsname\relax
  \providecommand{\doi}[1]{doi: #1}\else
  \providecommand{\doi}{doi: \begingroup \urlstyle{rm}\Url}\fi

\bibitem[Akbarzadeh-Sherbaf et~al.(2020)Akbarzadeh-Sherbaf, Safari, and
  Vahabie]{akbarzadeh2020digital}
Akbarzadeh-Sherbaf, K., Safari, S., and Vahabie, A.-H.
\newblock A digital hardware implementation of spiking neural networks with
  binary {FORCE} training.
\newblock \emph{Neurocomputing}, 412:\penalty0 129--142, 2020.

\bibitem[Akiba et~al.(2019)Akiba, Sano, Yanase, Ohta, and
  Koyama]{akiba2019optuna}
Akiba, T., Sano, S., Yanase, T., Ohta, T., and Koyama, M.
\newblock Optuna: A next-generation hyperparameter optimization framework.
\newblock In \emph{Proceedings of the 25th ACM SIGKDD International Conference
  on Knowledge Discovery \& Data Mining}, pp.\  2623--2631, 2019.

\bibitem[Amir et~al.(2017)]{amir2017low}
Amir, A. et~al.
\newblock A low power, fully event-based gesture recognition system.
\newblock In \emph{Proceedings of the IEEE Conference on Computer Vision and
  Pattern Recognition}, pp.\  7243--7252, 2017.

\bibitem[Azghadi et~al.(2020)Azghadi, Lammie, Eshraghian, Payvand, Donati,
  Linares-Barranco, and Indiveri]{azghadi2020hardware}
Azghadi, M.~R., Lammie, C., Eshraghian, J.~K., Payvand, M., Donati, E.,
  Linares-Barranco, B., and Indiveri, G.
\newblock Hardware implementation of deep network accelerators towards
  healthcare and biomedical applications.
\newblock \emph{IEEE Transactions on Biomedical Circuits and Systems},
  14\penalty0 (6):\penalty0 1138--1159, 2020.

\bibitem[Bellec et~al.(2018)Bellec, Salaj, Subramoney, Legenstein, and
  Maass]{bellec2018long}
Bellec, G., Salaj, D., Subramoney, A., Legenstein, R., and Maass, W.
\newblock Long short-term memory and learning-to-learn in networks of spiking
  neurons.
\newblock In \emph{Proceedings of the 32nd International Conference on Neural
  Information Processing Systems}, pp.\  795--805, 2018.

\bibitem[Bergstra et~al.(2011)Bergstra, Bardenet, Bengio, and
  K{\'e}gl]{bergstra2011algorithms}
Bergstra, J., Bardenet, R., Bengio, Y., and K{\'e}gl, B.
\newblock Algorithms for hyper-parameter optimization.
\newblock \emph{Advances in Neural Information Processing Systems}, 24, 2011.

\bibitem[Bienenstock et~al.(1982)Bienenstock, Cooper, and
  Munro]{bienenstock1982theory}
Bienenstock, E.~L., Cooper, L.~N., and Munro, P.~W.
\newblock Theory for the development of neuron selectivity: orientation
  specificity and binocular interaction in visual cortex.
\newblock \emph{Journal of Neuroscience}, 2\penalty0 (1):\penalty0 32--48,
  1982.

\bibitem[Bohte et~al.(2002)Bohte, Kok, and La~Poutre]{bohte2002error}
Bohte, S.~M., Kok, J.~N., and La~Poutre, H.
\newblock Error-backpropagation in temporally encoded networks of spiking
  neurons.
\newblock \emph{Neurocomputing}, 48\penalty0 (1-4):\penalty0 17--37, 2002.

\bibitem[Booij \& tat Nguyen(2005)Booij and tat Nguyen]{booij2005gradient}
Booij, O. and tat Nguyen, H.
\newblock A gradient descent rule for spiking neurons emitting multiple spikes.
\newblock \emph{Information Processing Letters}, 95\penalty0 (6):\penalty0
  552--558, 2005.

\bibitem[Courbariaux et~al.(2016)Courbariaux, Hubara, Soudry, El-Yaniv, and
  Bengio]{courbariaux2016binarized}
Courbariaux, M., Hubara, I., Soudry, D., El-Yaniv, R., and Bengio, Y.
\newblock Binarized neural networks: Training deep neural networks with weights
  and activations constrained to +1 or -1.
\newblock \emph{arXiv preprint arXiv:1602.02830}, 2016.

\bibitem[Cramer et~al.(2020)Cramer, Stradmann, Schemmel, and
  Zenke]{cramer2020heidelberg}
Cramer, B., Stradmann, Y., Schemmel, J., and Zenke, F.
\newblock The {Heidelberg} spiking data sets for the systematic evaluation of
  spiking neural networks.
\newblock \emph{IEEE Transactions on Neural Networks and Learning Systems},
  2020.

\bibitem[Diehl \& Cook(2015)Diehl and Cook]{diehl2015unsupervised}
Diehl, P.~U. and Cook, M.
\newblock Unsupervised learning of digit recognition using
  spike-timing-dependent plasticity.
\newblock \emph{Frontiers in Computational Neuroscience}, 9:\penalty0 99, 2015.

\bibitem[Eshraghian et~al.(2021)Eshraghian, Ward, Neftci, Wang, Lenz, Dwivedi,
  Bennamoun, Jeong, and Lu]{eshraghian2021training}
Eshraghian, J.~K., Ward, M., Neftci, E., Wang, X., Lenz, G., Dwivedi, G.,
  Bennamoun, M., Jeong, D.~S., and Lu, W.~D.
\newblock Training spiking neural networks using lessons from deep learning.
\newblock \emph{arXiv preprint arXiv:2109.12894}, 2021.

\bibitem[Feynman(1959)]{feynman1959plenty}
Feynman, R.~P.
\newblock Plenty of room at the bottom.
\newblock In \emph{APS Annual Meeting}, 1959.

\bibitem[Gerstner \& Kistler(2002)Gerstner and Kistler]{gerstner2002spiking}
Gerstner, W. and Kistler, W.~M.
\newblock \emph{Spiking neuron models: Single neurons, populations,
  plasticity}.
\newblock Cambridge University Press, 2002.

\bibitem[Hashemi et~al.(2018)Hashemi, Swersky, Smith, Ayers, Litz, Chang,
  Kozyrakis, and Ranganathan]{hashemi2018learning}
Hashemi, M., Swersky, K., Smith, J., Ayers, G., Litz, H., Chang, J., Kozyrakis,
  C., and Ranganathan, P.
\newblock Learning memory access patterns.
\newblock In \emph{International Conference on Machine Learning}, pp.\
  1919--1928. PMLR, 2018.

\bibitem[He et~al.(2019)He, Zhang, Zhang, Zhang, Xie, and Li]{he2019bag}
He, T., Zhang, Z., Zhang, H., Zhang, Z., Xie, J., and Li, M.
\newblock Bag of tricks for image classification with convolutional neural
  networks.
\newblock In \emph{Proceedings of the IEEE/CVF Conference on Computer Vision
  and Pattern Recognition}, pp.\  558--567, 2019.

\bibitem[Hubara et~al.(2016)Hubara, Courbariaux, Soudry, El-Yaniv, and
  Bengio]{hubara2016binarized}
Hubara, I., Courbariaux, M., Soudry, D., El-Yaniv, R., and Bengio, Y.
\newblock Binarized neural networks.
\newblock \emph{Advances in Neural Information Processing Systems}, 29, 2016.

\bibitem[Kang et~al.(2021)Kang, Choi, Eshraghian, Zhou, Kim, Kong, Zhu,
  Demirkol, Ascoli, Tetzlaff, et~al.]{kang2021build}
Kang, S.~M., Choi, D., Eshraghian, J.~K., Zhou, P., Kim, J., Kong, B.-S., Zhu,
  X., Demirkol, A.~S., Ascoli, A., Tetzlaff, R., et~al.
\newblock How to build a memristive integrate-and-fire model for spiking
  neuronal signal generation.
\newblock \emph{IEEE Transactions on Circuits and Systems I: Regular Papers},
  68\penalty0 (12):\penalty0 4837--4850, 2021.

\bibitem[Kheradpisheh et~al.(2021)Kheradpisheh, Mirsadeghi, and
  Masquelier]{kheradpisheh2021bs4nn}
Kheradpisheh, S.~R., Mirsadeghi, M., and Masquelier, T.
\newblock {BS4NN}: Binarized spiking neural networks with temporal coding and
  learning.
\newblock \emph{Neural Processing Letters}, pp.\  1--19, 2021.

\bibitem[Kingma \& Ba(2014)Kingma and Ba]{kingma2014adam}
Kingma, D.~P. and Ba, J.
\newblock Adam: A method for stochastic optimization.
\newblock \emph{arXiv preprint arXiv:1412.6980}, 2014.

\bibitem[Kozyrakis et~al.(2010)Kozyrakis, Kansal, Sankar, and
  Vaid]{kozyrakis2010server}
Kozyrakis, C., Kansal, A., Sankar, S., and Vaid, K.
\newblock Server engineering insights for large-scale online services.
\newblock \emph{IEEE Micro}, 30\penalty0 (4):\penalty0 8--19, 2010.

\bibitem[Laborieux et~al.(2021)Laborieux, Ernoult, Hirtzlin, and
  Querlioz]{laborieux2021synaptic}
Laborieux, A., Ernoult, M., Hirtzlin, T., and Querlioz, D.
\newblock Synaptic metaplasticity in binarized neural networks.
\newblock \emph{Nature Communications}, 12\penalty0 (1):\penalty0 1--12, 2021.

\bibitem[Lapique(1907)]{lapicque1907louis}
Lapique, L.
\newblock Recherches quantitatives sur l'excitation electrique des nerfs
  traitee comme une polarization.
\newblock \emph{Journal of Physiology and Pathology}, 9:\penalty0 620--635,
  1907.

\bibitem[LeCun et~al.(1998)LeCun, Bottou, Bengio, and
  Haffner]{lecun1998gradient}
LeCun, Y., Bottou, L., Bengio, Y., and Haffner, P.
\newblock Gradient-based learning applied to document recognition.
\newblock \emph{Proceedings of the IEEE}, 86\penalty0 (11):\penalty0
  2278--2324, 1998.

\bibitem[Loshchilov \& Hutter(2017)Loshchilov and Hutter]{loshchilov2017sgdr}
Loshchilov, I. and Hutter, F.
\newblock {SGDR}: Stochastic gradient descent with warm restarts.
\newblock \emph{International Conference on Learning Representations}, 2017.

\bibitem[Lu \& Sengupta(2020)Lu and Sengupta]{lu2020exploring}
Lu, S. and Sengupta, A.
\newblock Exploring the connection between binary and spiking neural networks.
\newblock \emph{Frontiers in Neuroscience}, 14:\penalty0 535, 2020.

\bibitem[Orchard et~al.(2021)Orchard, Frady, Rubin, Sanborn, Shrestha, Sommer,
  and Davies]{orchard2021efficient}
Orchard, G., Frady, E.~P., Rubin, D. B.~D., Sanborn, S., Shrestha, S.~B.,
  Sommer, F.~T., and Davies, M.
\newblock Efficient neuromorphic signal processing with loihi 2.
\newblock In \emph{2021 IEEE Workshop on Signal Processing Systems (SiPS)},
  pp.\  254--259. IEEE, 2021.

\bibitem[Paszke et~al.(2019)]{paszke2019pytorch}
Paszke, A. et~al.
\newblock Pytorch: An imperative style, high-performance deep learning library.
\newblock \emph{Advances in Neural Information Processing Systems},
  32:\penalty0 8026--8037, 2019.

\bibitem[Patrick et~al.(2008)Patrick, Posch, and Delbruck]{patrick2008128x}
Patrick, L., Posch, C., and Delbruck, T.
\newblock A 128$\times$128 120 db 15$\mu$ s latency asynchronous temporal
  contrast vision sensor.
\newblock \emph{IEEE Journal of Solid-State Circuits}, 43:\penalty0 566--576,
  2008.

\bibitem[Perez-Nieves \& Goodman(2021)Perez-Nieves and
  Goodman]{perez2021sparse}
Perez-Nieves, N. and Goodman, D.~F.
\newblock Sparse spiking gradient descent.
\newblock \emph{arXiv preprint arXiv:2105.08810}, 2021.

\bibitem[Perez-Nieves et~al.(2021)Perez-Nieves, Leung, Dragotti, and
  Goodman]{perez2021neural}
Perez-Nieves, N., Leung, V.~C., Dragotti, P.~L., and Goodman, D.~F.
\newblock Neural heterogeneity promotes robust learning.
\newblock \emph{Nature Communications}, 12\penalty0 (5791), 2021.

\bibitem[Rahimi~Azghadi et~al.(2020)Rahimi~Azghadi, Chen, Eshraghian, Chen,
  Lin, Amirsoleimani, Mehonic, Kenyon, Fowler, Lee,
  et~al.]{rahimi2020complementary}
Rahimi~Azghadi, M., Chen, Y.-C., Eshraghian, J.~K., Chen, J., Lin, C.-Y.,
  Amirsoleimani, A., Mehonic, A., Kenyon, A.~J., Fowler, B., Lee, J.~C., et~al.
\newblock Complementary metal-oxide semiconductor and memristive hardware for
  neuromorphic computing.
\newblock \emph{Advanced Intelligent Systems}, 2\penalty0 (5):\penalty0
  1900189, 2020.

\bibitem[Roy et~al.(2019)Roy, Jaiswal, and Panda]{roy2019towards}
Roy, K., Jaiswal, A., and Panda, P.
\newblock Towards spike-based machine intelligence with neuromorphic computing.
\newblock \emph{Nature}, 575\penalty0 (7784):\penalty0 607--617, 2019.

\bibitem[Rumelhart et~al.(1986)Rumelhart, Hinton, and
  Williams]{rumelhart1986learning}
Rumelhart, D.~E., Hinton, G.~E., and Williams, R.~J.
\newblock Learning representations by back-propagating errors.
\newblock \emph{Nature}, 323\penalty0 (6088):\penalty0 533--536, 1986.

\bibitem[Schaefer \& Joshi(2020)Schaefer and Joshi]{schaefer2020quantizing}
Schaefer, C.~J. and Joshi, S.
\newblock Quantizing spiking neural networks with integers.
\newblock In \emph{International Conference on Neuromorphic Systems 2020}, pp.\
   1--8, 2020.

\bibitem[Shaban et~al.(2021)Shaban, Bezugam, and Suri]{shaban2021adaptive}
Shaban, A., Bezugam, S.~S., and Suri, M.
\newblock An adaptive threshold neuron for recurrent spiking neural networks
  with nanodevice hardware implementation.
\newblock \emph{Nature Communications}, 12\penalty0 (1):\penalty0 1--11, 2021.

\bibitem[Shrestha \& Orchard(2018)Shrestha and Orchard]{shrestha2018slayer}
Shrestha, S.~B. and Orchard, G.
\newblock {SLAYER}: Spike layer error reassignment in time.
\newblock In \emph{Proceedings of the 32nd International Conference on Neural
  Information Processing Systems}, pp.\  1419--1428, 2018.

\bibitem[Voelker et~al.(2020)Voelker, Rasmussen, and
  Eliasmith]{voelker2020spike}
Voelker, A.~R., Rasmussen, D., and Eliasmith, C.
\newblock A spike in performance: Training hybrid-spiking neural networks with
  quantized activation functions.
\newblock \emph{arXiv preprint arXiv:2002.03553}, 2020.

\bibitem[Werbos(1990)]{werbos1990backpropagation}
Werbos, P.~J.
\newblock Backpropagation through time: What it does and how to do it.
\newblock \emph{Proceedings of the IEEE}, 78\penalty0 (10):\penalty0
  1550--1560, 1990.

\bibitem[Xiao et~al.(2017)Xiao, Rasul, and Vollgraf]{xiao2017fashion}
Xiao, H., Rasul, K., and Vollgraf, R.
\newblock Fashion-{MNIST}: A novel image dataset for benchmarking machine
  learning algorithms.
\newblock \emph{arXiv preprint arXiv:1708.07747}, 2017.

\bibitem[Xu et~al.(2013)Xu, Zeng, Han, and Yang]{xu2013supervised}
Xu, Y., Zeng, X., Han, L., and Yang, J.
\newblock A supervised multi-spike learning algorithm based on gradient descent
  for spiking neural networks.
\newblock \emph{Neural Networks}, 43:\penalty0 99--113, 2013.

\bibitem[Zenke \& Vogels(2021)Zenke and Vogels]{zenke2021remarkable}
Zenke, F. and Vogels, T.~P.
\newblock The remarkable robustness of surrogate gradient learning for
  instilling complex function in spiking neural networks.
\newblock \emph{Neural Computation}, 33\penalty0 (4):\penalty0 899--925, 2021.

\bibitem[Zhang \& Linden(2003)Zhang and Linden]{zhang2003other}
Zhang, W. and Linden, D.~J.
\newblock The other side of the engram: experience-driven changes in neuronal
  intrinsic excitability.
\newblock \emph{Nature Reviews Neuroscience}, 4\penalty0 (11):\penalty0
  885--900, 2003.

\end{thebibliography}
\bibliographystyle{icml2022}


\newpage
\appendix
\onecolumn
\section{Spiking Neuron Model}\label{app:a1}


The dynamics of the passive membrane modeled using an RC circuit can be represented as:

\begin{equation}\label{aeq:ode}
    \tau_m\frac{du_t}{dt} = -u_t + i_tr
\end{equation}

where $u_t$ is the hidden state (membrane potential) of the neuron at time $t$, $\tau_m=rc$ is the time constant of the membrane, $r$ and $c$ are the resistance and capacitance of the membrane, respectively, and $i_t$ is the input current injection at time $t$. 

The goal is to convert \Cref{aeq:ode} into a recurrent representation with fewer hyperparameters to enable training using backprop through time. The forward Euler method is used to find an approximate solution to \Cref{aeq:ode} without taking the limit $\Delta t \rightarrow 0$:

\begin{equation}\label{aeq:euler}
    \tau_m \frac{u_{t+\Delta t} - u_t}{\Delta t} = -u_t + i_tr
\end{equation}

For sufficiently small $\Delta t$, this provides a reasonable approximation of continuous-time integration. Isolating $u_{t+\Delta t}$ gives:

\begin{equation}\label{aeq:eul2}    
    u_{t+\Delta t} = (1-\frac{\Delta t}{\tau_m})u_t + \frac{\Delta t}{\tau_m}i_tr
\end{equation}

The change of $u_t$ from one time step to the next can be found in absence of input current: 

\begin{equation}\label{aeq:beta}
    \beta = \frac{u_{t+\Delta t}}{u_t} = (1- \frac{\Delta t}{\tau_m})
\end{equation}

To reduce the number of free variables in \Cref{aeq:eul2}, normalize $\Delta t=1$, and let $r=1 \Omega$: 

\begin{equation} \label{aeq:2.2}
    u_{t+1} = \beta u_t + (1-\beta)i_t
\end{equation}


To further simplify the model, let the effect of $(1-\beta)i_t$ be approximated by a learnable weight scaled by the output of the previous layer, $(1-\beta)i_t \leftarrow \sum_i w^{ij}z^i_t$. For computational efficiency (i.e., to enable inputs to be directly propagated from the input to the output of the network without delay), a time-shift is applied to $z_t^i$:

\begin{equation} \label{aeq:2.3}
    u^j_{t+1} = \beta u_t^j + \sum_i w^{ij}z_{t+1}^i
\end{equation}

Finally, the term $-z_t^j \theta$ is added to \Cref{aeq:2.3} to account for the neuronal reset mechanism where the threshold is subtracted from the membrane potential each time a spike $z_t^j=1$ occurs:

\begin{equation} \label{aeq:2.4}
    u^j_{t+1} = \beta u_t^j + \sum_i w^{ij}z_{t+1}^i - \theta z_{t}^j.
\end{equation}

\newpage

\section{Threshold Annealing Algorithm}\label{app:alg}

\begin{algorithm*}[h]
   \caption{Threshold Annealing: Same as \Cref{alg:tha}, but with temporal iteration decoupled from minibatch iteration}
   \label{alg:tha2}
\begin{algorithmic}
   \REQUIRE $\theta_0 \in \mathbb{R}_{>0}$: Initial threshold
   \REQUIRE $\theta_\infty \in \mathbb{R}_{>\theta_0}$: Final threshold
   \REQUIRE $\alpha_\theta \in \mathbb{R}_{>0}:$ Inverse threshold time constant
   \REQUIRE $w:$ Initial parameters
   \REQUIRE $\eta:$ Learning rate
   \STATE $\theta_\gamma \leftarrow \theta_0$ (Initialize threshold)
   \STATE $s \leftarrow \partial \Tilde{z}(\theta_0)/\partial u$ (Initialize surrogate gradient using $\theta_0$)
   \STATE $\gamma \leftarrow 0$ (Initialize training step)
   \WHILE {$w$ not converged}
   \STATE $\gamma \leftarrow \gamma + 1$
   \FOR{$t\in\{1,...,N\}$}
  \STATE $\mathcal{L} \leftarrow f(w_{\gamma - 1}, \theta_\gamma)$ (Get loss by iterating through $N$ time-steps during forward-pass)
  \STATE $g_\gamma \leftarrow \nabla_w \mathcal{L}$ (Get gradients w.r.t. surrogate gradient $s$ using initial threshold $\theta_0$)
   \ENDFOR
   \STATE $w_\gamma \leftarrow w_{\gamma - 1} - \eta \cdot g_\gamma$ (Update parameters e.g. using SGD or Adam)
   \STATE $\theta_\gamma \leftarrow \theta_{\gamma-1} + \alpha(\theta_\infty - \theta_{\gamma-1})$ (Update threshold \textit{outside} of forward-pass temporal loop, without updating $s$)
    \ENDWHILE
    \STATE \textbf{return: }$w_\gamma$ (Resulting parameters) 
\end{algorithmic}
\end{algorithm*}

\Cref{alg:tha2} is a more detailed form of \Cref{alg:tha} to highlight how the temporal-loop during the forward-pass is decoupled from the minibatch training loop. Homeostasis would place the threshold update $\theta_\gamma$ within the for-loop to stabilize neuronal activity, whereas threshold annealing takes it out to increase the effective hidden state precision whilst reducing both memory and computational complexity.

\section{Upper Bound on the Rate of Threshold Annealing}\label{app:a3}


To minimize the impact threshold updates have on a network's ability to learn, we set the constraint that the threshold update cannot exceed the weight update-induced membrane potential change. Formally, the threshold update $\Delta \theta_\gamma = \theta_{\gamma + 1} - \theta_{\gamma}$ is bounded by how far the new membrane potential at $\gamma+1$ is from the original threshold at $\gamma$:

\begin{equation} \label{aeq:c1}
    \Delta \theta_\gamma < u_{\gamma + 1} - \theta_\gamma
\end{equation}

For simplicity, we assume this takes place at the first spike time for an instantaneous input (which is why notation for time $t$ has been dropped), though this constraint will be relaxed. $\Delta \theta_\gamma = \theta_{\gamma+1} - \theta_\gamma$ can be found from \Cref{eq:tha2}:

\begin{equation}\label{aeq:c2}
    \Delta \theta_\gamma = \alpha(\theta_\infty - \theta_\gamma)
\end{equation}

which is substituted into \Cref{aeq:c1}:

\begin{equation}\label{aeq:c3}
    \alpha (\theta_\infty - \theta_\gamma) < u_{\gamma+1} - \theta_\gamma
\end{equation}


Considering only one spiking neuron, the immediate post-synaptic current influence at the spike time from \Cref{eq:lif} for iterations $\gamma$ and $\gamma+1$ is given below:

\begin{equation}\label{aeq:c4}
    u_{\gamma} = \mathbf{z} \cdot \mathbf{w}_{\gamma}^{~:, j}
\end{equation}

\begin{equation}\label{aeq:c5}
    u_{\gamma + 1} = \mathbf{z} \cdot \mathbf{w}_{\gamma+1}^{~:, j}
\end{equation}

where $\mathbf{z} \in \mathbb{R}^i$ is the input vector, $\mathbf{w}_{\gamma}^{~:, j} \in \mathbb{R}^i$ is the weight vector connected to the $j^{th}$ neuron of the first layer (and can be generalized to subsequent layers). For simplicity, assume the weight update is calculated via stochastic gradient descent (SGD), though noting that the final term below should be substituted for the preferred optimization update rule:

\begin{equation}\label{aeq:c6}
    \mathbf{w}^{~:, j}_{\gamma + 1} = \mathbf{w}_{\gamma}^{~:, j} + \eta \frac{\partial \mathcal {L}}{\partial \mathbf{w}^{~:,j}_\gamma}
\end{equation}

\Cref{aeq:c6} is substituted back into \Cref{aeq:c5}:

\begin{equation}\label{aeq:c7}
     u_{\gamma + 1} = \mathbf{z} \cdot (\mathbf{w}_{\gamma}^{~:, j} + \eta \frac{\partial \mathcal {L}}{\partial \mathbf{w}^{~:,j}_\gamma})
\end{equation}

which is then substituted into \Cref{aeq:c3}:

\begin{equation}\label{aeq:c8}
    \alpha (\theta_\infty - \theta_\gamma)  < \mathbf{z} \cdot (\mathbf{w}_{\gamma}^{~:, j} + \eta \frac{\partial \mathcal {L}}{\partial \mathbf{w}^{~:,j}_\gamma}) - \theta_\gamma
\end{equation}

\begin{equation}\label{aeq:c9}
    \implies \underbrace{(1-\alpha)}_{\rm const.}\theta_\gamma + \underbrace{\alpha\theta_\infty}_{\rm const.} < \underbrace{\mathbf{z}}_{\rm const.} \cdot~ \Big(\mathbf{w}_{\gamma}^{~:, j} + \underbrace{\eta \frac{\partial \mathcal {L}}{\partial \mathbf{w}^{~:,j}_\gamma}}_{\rm update}\Big)
\end{equation}

All constant terms have been underlined, and the evolution of $\mathbf{w}^{~:,j}_\gamma$ is dependent on $\partial \mathcal{L}/\partial \mathbf{w}^{~:,j}_\gamma$. When using more common optimizers, such as Adam or SGD with momentum, the weight update term must be swapped out. For the Adam optimizer, the update term will follow an exponential decay as the network approaches training convergence until the right hand side of \Cref{aeq:c9} stabilizes at $\mathbf{z} \cdot \mathbf{w}_{\gamma}^{~:, j}$. The only reasonable and valid solution as $\gamma \rightarrow \infty$ is for $\theta_\gamma$ to exponentially relax to $\theta_\infty$, where the update term approaches zero where adaptive/exponential decay is applied, and the term on the left factors out to:

\begin{equation}\label{aeq:c10}
        \theta_\infty < \mathbf{z} \cdot~ \mathbf{w}_{\infty}^{~:, j}
\end{equation}

$\theta_\gamma$ and $\theta_\infty$ taking on values much less than the dot product of the input and weights would be a mathematically valid solution, but at the cost of each neuron no longer having a sufficiently wide state-space and the ability to trigger spikes (see \Cref{sec:BSNNs} for detail on why memory and spike emission become mutually exclusive as the threshold is reduced). Therefore, an exponential relaxation to a steady-state threshold described in \Cref{eq:tha1} is the optimal solution for exponentially adaptive weight decays.

Furthermore, $\mathbf{z}$ is a time-varying input (it is only constant across iterations given it represents the same input data), and so \Cref{aeq:c10} must be generalized to account for the historical contributions of post-synaptic current to the membrane state at the spike time (\Cref{eq:lif}).

The above conditions were derived assuming that the threshold update should not impact spiking activity at all, which is unnecessarily stringent. As long as each neuron fires at least once, then the immediate weights attached to its input will contribute to the output loss, have a gradient signal (for dense layers), and therefore avoid becoming dead neurons. A more reasonable objective would to be to only introduce the constraint from \Cref{aeq:c10} when a neuron is emitting at most a single spike across the simulation run time.

\newpage

\section{Experimental Results} \label{app:exp}

\textbf{Parameter Initialization: } For all experiments, parameters were initialized using the default methods in PyTorch 1.10.1.  Dense layers were initialized by uniformly sampling from \textit{U}($-\sqrt{a}$, $\sqrt{a}$):

\begin{equation}\label{eq:initlin}
    a = \frac{1}{N_{\rm in}}, 
\end{equation}

\noindent where $N_{\rm in}$ is the number of input features. Convolutional layer parameters were also uniformly sampled:

\begin{equation}
    a = \frac{1}{C_{\rm in}N_x N_y}
\end{equation}

\noindent where $C_{\rm in}$ is the number of input channels, and $N_x$ and $N_y$ are the kernel dimensions.

\textbf{Hyperparameter search: } 
The following hyperparameters were searched in all experiments:

\begin{itemize}
    \item $\beta$: neuron decay rate/membrane potential inverse time constant (\Cref{eq:lif})
    \item $\theta$: firing threshold (\Cref{eq:spk})
    \item $k$: surrogate gradient slope (\Cref{eq:surrogate})
    \item $\eta$: initial learning rate
    \item \textbf{GC}: gradient clipping 
    \item \textbf{WC}: weight clipping
    \item \textbf{DO}: dropout rate (dense layers)
    \item \textbf{BN}: batch normalization (convolutional layers)
\end{itemize}

The following hyperparameters were searched in all threshold annealing experiments:

\begin{itemize}
    \item $\theta_0^L$: initial firing threshold of all neurons in layer $L$ (\Cref{eq:tha2})
    \item $\theta_\infty^L$: final firing threshold of all neurons in layer $L$ (\Cref{eq:tha2})
    \item $\alpha^L$: inverse time constant of threshold of all neurons in layer $L$ (\Cref{eq:tha2})
\end{itemize}

Hyperparameters were selected based on 500 separate trials for each experiment using a tree-structured Parzen Estimator algorithm to randomly sample from the search space in Optuna \cite{akiba2019optuna, bergstra2011algorithms}. Based on this approach, individual hyperparameters were found for high precision, BSNN, BSNN with normalized threshold, and BSNN with threshold annealing across all four datasets. I.e., 16 different sets of hyperparameters were found.

With respect to the threshold:

\begin{itemize}
    \item High precision and BSNN experiments set the threshold as a free hyperparameter,
    \item BSNN with normalized threshold fixes the threshold at $\theta=1$, and
    \item BSNN with threshold annealing set the initial threshold $\theta_0$, final threshold $\theta_\infty$, and inverse time constant $\alpha$ as free hyperparameters. 
\end{itemize}

Due to the additional hyperparameters in the BSNN with threshold annealing experiment, the final result would likely benefit from using additional trials during the hyperparameter search. But for fairness, this was fixed to 500 trials as with all other cases.

\textbf{Optimization: } Unless otherwise specified, the Adam optimizer was used with 1st and 2nd order moment estimates set to ($\beta_1, \beta_2$) = ($0.9, 0.999$). A cosine annealed learning rate with warm restarts is computed using the following scheduler~\cite{loshchilov2017sgdr, he2019bag}:

\begin{equation}
    \eta_t = \frac{1}{2} \eta \Big(1 + \text{cos}\Big(\frac{\pi \gamma}{T} \Big) \Big),
\end{equation}

\noindent where $\eta$ is the initial learning rate, $\gamma$ is the iteration, and $T$ is the period of the schedule. For all cases where cosine annealing is used, $T$ is set to a period of 10 training epochs.

The fast sigmoid surrogate gradient was used for all spiking neurons (\Cref{eq:surrogate}).


\subsection{Temporal Coding} \label{app:timing}

A corresponding notebook to replicate these experiments, in addition to animated visualizations of the threshold evolution and its impact on membrane potential, are provided in the source code.

\textbf{Architecture: } A 3-layer fully connected network was used with 100$-$1000$-$1 neurons.

\textbf{Hyperparameters: } The hyperparameters of the temporal coding task from \Cref{sec:TC} are shown in the table below, and were chosen arbitrarily for the sake of an intuitive demonstration of threshold annealing. 

 \begin{table*}[!ht]\centering 
\caption{Hyperparameters for Temporal Coding Task}\label{tab:temp_params}
\vskip 0.15in
\begin{threeparttable}
\begin{tabular}{cccccccc} \toprule \toprule
\textbf{Precision}  & \textbf{$\beta$} & \textbf{$\theta$} & \textbf{$\theta_0$} &\textbf{$\theta_\infty$} & \textbf{$\alpha$} &\textbf{$k$} & \textbf{$\eta$} \\ \midrule
flt32 & 0.6 & 2.0 & $-$ & $-$ &  $-$ &  5.0 & 1e-3 \\
BSNN; $\theta$=1 &  0.15 & 1.0 & $-$ & $-$ &  $-$ &  5.0 & 1e-3 \\
BSNN; $\theta$=50 &  0.15 & 50.0 & $-$ & $-$ &  $-$ &  5.0 & 1e-3 \\
Proposed &  0.15 & $-$ & 5.0 & 50.0 &  5e-3 & 5.0 & 1e-3 \\
\bottomrule \bottomrule

\end{tabular}
 \begin{tablenotes}
    \item[1] Stochastic Gradient Descent with momentum of 0.9 was used for all temporal coding experiments. No schedule was used.
  \end{tablenotes}
 \end{threeparttable}
\end{table*}

\textbf{Loss: } Let $y_t$ be the target membrane potential of the output neuron at time $t$. The total mean square error is calculated by summing the loss across all time steps $T$:

\begin{equation}\label{beq:15}
    \mathcal{L}_{MSE} =\sum_t^T(y_t-u_t)^2.
\end{equation}

\textbf{Peak membrane target: } The peak of the target membrane potential in \Cref{fig:time} is set to be 10\% above the threshold to increase the probability of spike emission.

\newpage 

\subsection{Static Datasets}\label{app:static}

\textbf{Hyperparameters: } The hyperparameters used in the static dataset experiments from \Cref{sec:static} are shown in the table below.

\begin{table*}[!ht]\centering 
\caption{Hyperparameters for Static Datasets}\label{tab:params}
\vskip 0.15in
\begin{threeparttable}
\begin{tabular}{cccccccccc} \toprule \toprule
\textbf{Dataset} & \textbf{Precision}  & \textbf{$\beta$} & \textbf{$\theta$} & \textbf{$k$} & \textbf{$\eta$} & \textbf{GC} & \textbf{WC} & \textbf{BN} & \textbf{DO} \\ \midrule
MNIST & flt32 & 0.92 & 2.0 & 6.0 & 1.9e-3 & \ding{51} & \ding{51} & \ding{51} & 0.09\\
MNIST & BSNN;$\theta$:1 &  0.74 & 1.0 & 4.7 & 6.5e-3 & \ding{55} & \ding{55} & \ding{51} & 0.87 \\ 
MNIST & BSNN &  0.84 & 6.4 & 2.8 & 2.7e-3 & \ding{55} & \ding{55} & \ding{51} & 0.06 \\ 
MNIST & Proposed &  0.99 & $-$ & 10.2 & 1.0e-2 & \ding{55} & \ding{55} & \ding{51} & 0.03 \\ \midrule
FMNIST & flt32 & 0.39 & 1.5 & 7.7 & 2.0e-3 & \ding{51} & \ding{51} & \ding{51} & 0.13\\
FMNIST & BSNN;$\theta$:1 &  0.82 & 1.0 & 4.3 & 1.2e-3 & \ding{51} & \ding{51} & \ding{51} & 0.86 \\ 
FMNIST & BSNN &  0.87 & 9.7 & 6.7 & 1.6e-2 & \ding{51} & \ding{55} & \ding{51} & 0.01 \\ 
FMNIST & Proposed &  0.87 & $-$ & 0.2 & 8.4e-4 & \ding{55} & \ding{55} & \ding{51} & 0.65 \\ \midrule
\bottomrule \bottomrule

\end{tabular}
 \begin{tablenotes}
    \item[1] 16Conv5-AP2-64Conv5-AP2-1024Dense10.
    \item[2] Batch size of 128 used for all experiments.
    \item[3] MNIST flt32 did not use a learning rate schedule.
    \item[4] FMNIST Proposed used SGD with momentum set to 0.86.
  \end{tablenotes}
 \end{threeparttable}
\end{table*}

\begin{table*}[!ht]\centering 
\caption{Threshold Annealing Hyperparameters for Static Datasets}\label{tab:params}
\vskip 0.15in
\begin{threeparttable}
\begin{tabular}{cccccccccc} \toprule \toprule
\textbf{Dataset} & \textbf{$\theta_0^1$} & \textbf{$\theta_\infty^1$} & \textbf{$\alpha^1$} & \textbf{$\theta_0^2$} & \textbf{$\theta_\infty^2$} & \textbf{$\alpha^2$} & \textbf{$\theta_0^3$} & \textbf{$\theta_\infty^3$} & \textbf{$\alpha^3$} \\ \midrule

MNIST & 11.7 & 16.0 & 2.4e-2 & 14.1 & 30.4 & 0.12 & 0.7 & 4.2 & 1.1e-3 \\ \midrule

FMNIST & 6.9 & 14.0 & 3.7e-2 & 10.3 & 23.1 & 0.30 & 18.0 & 27.9 & 0.10  \\ 
\bottomrule \bottomrule

\end{tabular}
 \begin{tablenotes}
    \item[1] $\theta_0^L$ is the initial threshold of all neurons in layer $L$.
    \item[2] $\theta_\infty^L$ is the final threshold of all neurons in layer $L$.
    \item[3] $\alpha^L$ is the inverse time constant of the threshold for all neurons in layer $L$.
  \end{tablenotes}
 \end{threeparttable}
\end{table*}

\textbf{Loss: } To emulate the limitations of efficient neuromorphic hardware, we assume the membrane potential is an inaccessible hidden state when calculating the loss function \cite{azghadi2020hardware}. Instead, we train our network under the more challenging constraint of using target spike counts for each class. The spikes of each output neuron $z^j_t$ are accumulated over time. The mean square error loss of the target count $c^j$ is calculated and then summed across the $N$ output classes ($N=10$).

Let $y_t$ be the target membrane potential of the output neuron at time $t$. The total mean square error is calculated by summing the loss across all time steps $T$. We follow a similar approach to SLAYER \cite{shrestha2018slayer} where the target of the correct class $c$ is to fire 80\% of the time, and incorrect classes are set to fire 20\% of the time to avoid excessive suppression of neuronal activity:

\begin{equation}\label{eq:mse_count_loss}
    \mathcal{L}_{MSE} = \sum_j^{N}\sum_t(c^j - z^j_t)^2.
\end{equation}

\newpage
\subsection{Temporal Datasets} \label{app:temporal}
\textbf{Hyperparameters: } The hyperparameters used in the temporal dataset experiments from \Cref{sec:temporal} are shown in the table below.

\begin{table*}[!ht]\centering 
\caption{Hyperparameters for Temporal Datasets}\label{tab:params}
\vskip 0.15in
\begin{threeparttable}
\begin{tabular}{ccccccccccc} \toprule \toprule
\textbf{Dataset} & \textbf{Precision}  & \textbf{$\beta$} & \textbf{$\theta$} & \textbf{$k$} & \textbf{$\eta$} & \textbf{GC} & \textbf{WC} & \textbf{BN} & \textbf{DO$^1$} & \textbf{DO$^2$} \\ \midrule
DVS128 & flt32 & 0.72 & 2.5 & 5.4 & 7.5e-3 & \ding{55} & \ding{51} & \ding{51} & 0.29 & $-$\\
DVS128 & BSNN;$\theta$:1 &  0.1 & 1.0 & 0.1 & 3.6e-3 & \ding{55} & \ding{55} & \ding{55} & 0.64 & $-$ \\ 
DVS128 & BSNN &  0.78 & 8.7 & 0.3 & 9.6e-4 & \ding{51} & \ding{51} & \ding{55} & 0.06 & $-$ \\ 
DVS128 & Proposed &  0.93 & $-$ & 0.2 & 1.8e-3 & \ding{51} & \ding{55} & \ding{55} & 0.43 & $-$ \\ \midrule
SHD & flt32 & 0.98 & 1.7 & 2.0 & 2.7e-4 & \ding{55} & \ding{55} & $-$ & 0.05 & 0.04 \\
SHD & BSNN;$\theta$:1 &  0.61 & 1.0 & 0.4 & 1.9e-4 & \ding{51} & \ding{51} & $-$ & 0.09 & 0.11 \\ 
SHD & BSNN &  0.33 & 3.4 & 0.1 & 5.0e-4 & \ding{51} & \ding{51} & $-$ & 0.22 & 0.07 \\ 
SHD & Proposed & 0.95 & $-$ & 0.3 & 6.5e-4 & \ding{51} & \ding{51} & $-$ & 0.02 & 0.19 \\ \midrule
\bottomrule \bottomrule

\end{tabular}
 \begin{tablenotes}
    \item[1] DVS128: 16Conv5-AP2-32Conv5-AP2-800Dense11
    \item[2] DVS128 flt32 batch size: 16. DVS128 BSNN batch size: 8.
    \item[3] DVS128 events integrated over 5~ms.
    \item[4] SHD: 700Dense3000Dense20
    \item[5] SHD batch size: 32
    \item[6] SHD flt32 events integrated over 2~ms; BSNN events integrated over 3~ms.
    \item[7] DO$^1$: dropout rate in the first dense layer, DO$^2$: dropout rate in second dense layer
  \end{tablenotes}
 \end{threeparttable}
\end{table*}

\begin{table*}[!ht]\centering 
\caption{Threshold Annealing Hyperparameters for Temporal Datasets}\label{tab:temporal}
\vskip 0.15in
\begin{threeparttable}
\begin{tabular}{cccccccccc} \toprule \toprule
\textbf{Dataset} & \textbf{$\theta_0^1$} & \textbf{$\theta_\infty^1$} & \textbf{$\alpha^1$} & \textbf{$\theta_0^2$} & \textbf{$\theta_\infty^2$} & \textbf{$\alpha^2$} & \textbf{$\theta_0^3$} & \textbf{$\theta_\infty^3$} & \textbf{$\alpha^3$} \\ \midrule

DVS128 Gesture & 10.4 & 12.2 & 3.3e-3 & 16.6 & 19.1 & 6.1e-3 & 6.8 & 16.5 & 0.17 \\ \midrule

SHD & 13.5 & 45.2 & 2.8e-5 & 11.2 & 51.1 & 1.36e-5 & $-$ & $-$ & $-$  \\ 
\bottomrule \bottomrule

\end{tabular}
 \begin{tablenotes}
    \item[1] $\theta_0^L$ is the initial threshold of all neurons in layer $L$.
    \item[2] $\theta_\infty^L$ is the final threshold of all neurons in layer $L$.
    \item[3] $\alpha^L$ is the inverse time constant of the threshold for all neurons in layer $L$.
  \end{tablenotes}
 \end{threeparttable}
\end{table*}

\textbf{Loss: } For both networks, the same loss function is used as in \Cref{eq:mse_count_loss}.

\newpage 

\section{Discussion}\label{app:dis}
\subsection{Spike Activity on Static Datasets} \label{app:dist}

\begin{figure*}[!h] 
\centering
\subfloat[]
{
	\includegraphics[scale=0.3]{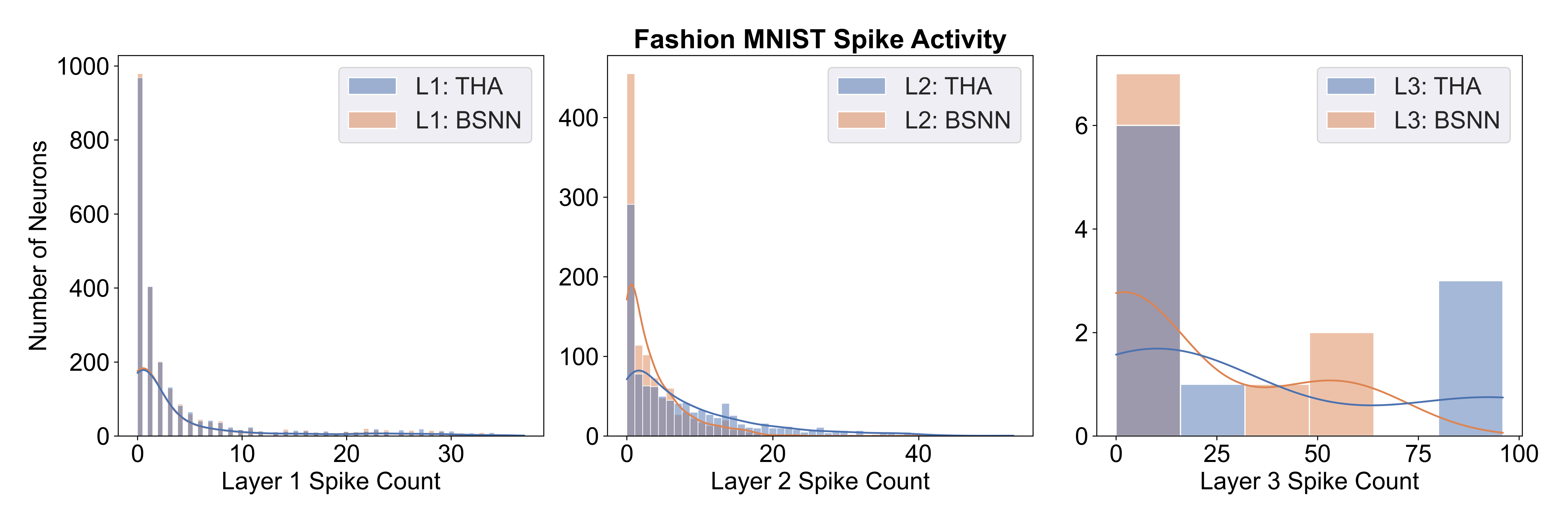}
	\label{fig1a}
}\\
\subfloat[]
{
	\raisebox{3.8em}{\includegraphics[scale=0.3]{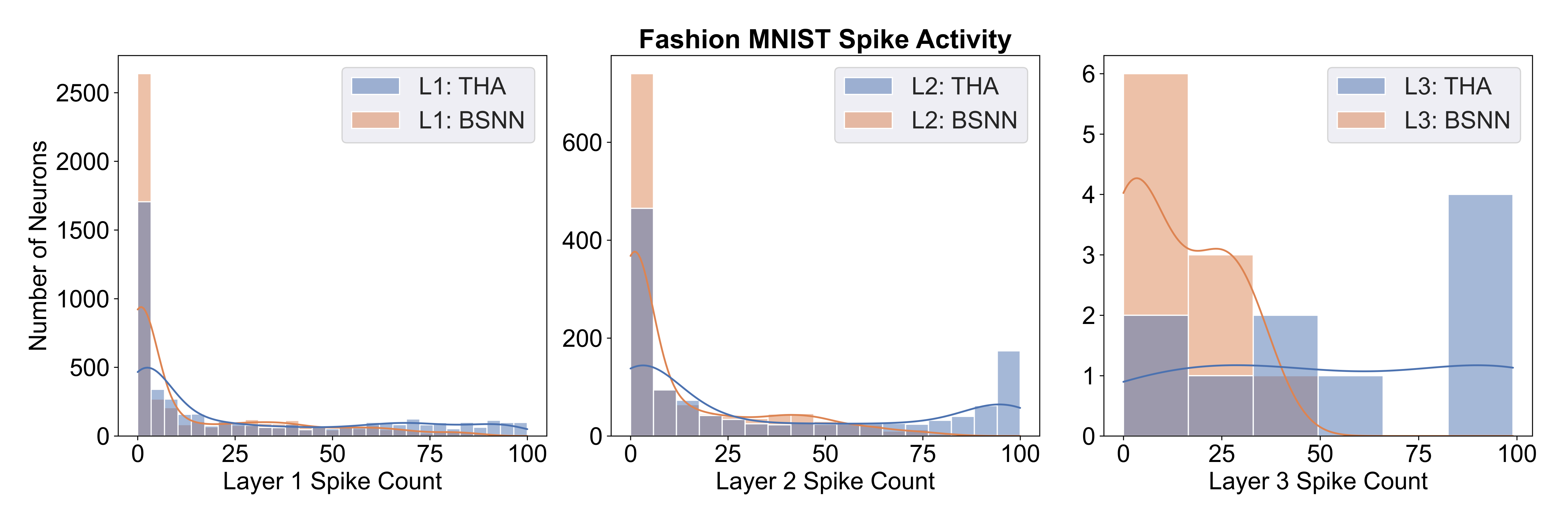}}
	\label{fig1b}
}
\caption{Spiking activity across layers at the start of the training process of BSNNs with threshold annealing (THA) and without (BSNN) on static datasets.}
\label{figapp:static}
\end{figure*}

\end{document}